\RequirePackage{fix-cm}

\documentclass[smallextended]{svjour3}

\smartqed 

\usepackage{cite}
\usepackage{amsmath,amssymb,amsfonts}
\usepackage{algorithmic}
\usepackage{graphicx}
\usepackage{xcolor}
\def\BibTeX{{\rm B\kern-.05em{\sc i\kern-.025em b}\kern-.08em
    T\kern-.1667em\lower.7ex\hbox{E}\kern-.125emX}}
\usepackage{lineno}
\usepackage{hyperref}
\usepackage{float}
\usepackage{multirow}
\usepackage{caption}
\usepackage{subcaption}
\usepackage{mathdots}
\usepackage[classicReIm]{kpfonts}
\usepackage[english]{babel} 
\usepackage{enumitem}
\usepackage[section]{placeins}
\usepackage{wrapfig}

\begin{document}

\title{Computational Intelligence Approach to Improve the Classification Accuracy of Brain Neoplasm in MRI Data
}

\author{Nilanjan Sinhababu         \and
        Monalisa Sarma          \and
        Debasis Samanta
}

\titlerunning{Detection of Brain Neoplasm in MRI Data}
\authorrunning{Sinhababu et al.} 

\institute{N. Sinhababu \at
              Subir Chowdhury School of Quality and Reliability\\
              IIT Kharagpur\\
              Mobile: +91-8617826402\\
              \email{nilanjansb95@gmail.com}           
           \and
           M. Sarma \at
              Subir Chowdhury School of Quality and Reliability\\
              IIT Kharagpur\\
            \and
            D.Samanta \at
              Department of Computer Science and Engineering\\
              IIT Kharagpur\\
}

\maketitle

\begin{abstract}
Automatic detection of brain neoplasm in Magnetic Resonance Imaging (MRI) is gaining importance in many medical diagnostic applications. This report presents two improvements for brain neoplasm detection in MRI data: an advanced preprocessing technique is proposed to improve the area of interest in MRI data and a hybrid technique using Convolutional Neural Network (CNN) for feature extraction followed by Support Vector Machine (SVM) for classification. The learning algorithm for SVM is modified with the addition of cost function to minimize false positive prediction addressing the errors in MRI data diagnosis. The proposed approach can effectively detect the presence of neoplasm and also predict whether it is cancerous (malignant) or non-cancerous (benign). To check the effectiveness of the proposed preprocessing technique, it is inspected visually and evaluated using training performance metrics. A comparison study between the proposed classification technique and the existing techniques was performed. The result showed that the proposed approach outperformed in terms of accuracy and can handle errors in classification better than the existing approaches.

\keywords{Medical imaging\and Classification of brain tumor data\and Brain neoplasm diagnosis\and MRI data classification}

\end{abstract}

\section{Introduction}
\label{S:1}

Brain neoplasm is known as the deadliest of all forms of
cancer. Any abnormalities in the growth of neural cells
within the brain skull or the spinal cord lead to a brain
neoplasm or more generally known as a brain tumor.
Around two-thirds of adults are diagnosed with aggressive brain cancer resulting in death within 2 years of diagnosis \cite{chinot2014bevacizumab, gilbert2014randomized}. The primary cause of death due to health-related issues related to brain neoplasm is due to the unavailability of any special cure after a certain stage \cite{aldape2019challenges}. It is worth to mention that any kind of neoplasm must be detected as early as possible.
Diagnosis of any brain tumor is done by a neurological test, for example, CT scan, Magnetic Resonance Imaging (MRI)\cite{osborn2015diagnostic,armstrong2004imaging}, etc. MRI is one of the advanced medical imaging techniques and is used to produce high-quality images \cite{liang2000principles} to study tumors in soft tissues of the brain. The need for a computationally intelligent approach for the classification of brain neoplasm is required for the following reasons:
\begin{itemize}
\item MRI provides the location and size of a tumor efficiently but has no way of classifying its type \cite{aronen1994cerebral, young2006brain}.
\item For classification of the MRI, a biopsy is required, this process is painful, time-consuming, and has limitations such as sampling error and variability in interpretation \cite{aronen1994cerebral, de1997proliferative}.
\item Visual inspection and detection may result in misdiagnosis due to human errors caused by visual fatigue \cite{de1997proliferative}.
\item Daily reporting of a large number of new cases implies a high chance of misdiagnosis due to the time limitations, interaction and doctor to patient ratio \cite{ford1996doctor, bagcchi2015india}.
\item It is necessary to aid the medical fraternity to understand the tumor type automatically and hence helps in reducing load and saving time.
\item Computerised analysis proven to be more superior and are being researched for predicting cancerous cell in a brain MRI data.
\end{itemize}

To achieve the goal of developing an automatic brain neoplasm detection system, various techniques are used in the literature. The three most important stages present in these systems are data preprocessing, feature optimization, and classification. Existing preprocessing techniques consider noise removal while preserving texture information. Image preprocessing techniques, such as cropping, size normalization, histogram equalization based contrast enhancement \cite{amin2012brain}, Gaussian filter-based noise removal \cite{mahajanidetection}, and Center Weighted Median(CWM) \cite{george2012mri} based techniques are used. The Principal Component Analysis (PCA) \cite{amin2012brain, othman2011probabilistic, singh2012classification} and Gray Level Co-occurrence Matrix (GLCM) \cite{joshi2010classification, mahajanidetection} based techniques are used in majority of the works for feature extraction. Some of the works \cite{kharrat2010hybrid} also used Spatial Gray Level Dependence Method (SGLDM) to extract wavelet-based textures. Some of the methodologies used manual intervention to extract texture and intensity features \cite{jafari2012hybrid}. For classification, the majority of the work use Multi-Layer Perceptron (MLP), Probabilistic Neural Network (PNN), Support Vector Machine(SVM), etc. \cite{amin2012brain, singh2012classification, othman2011probabilistic, kharrat2010hybrid}. There exist works \cite{jafari2012hybrid, mahajanidetection}, which have combined multiple classifiers to improve the classification accuracy, such as Back Propagation Neural Network(BPNN) and K-Nearest Neighbours(KNN). Since, stacked MRI image slices can be used as a 3D representation of the MRI data, few techniques use 3D-CNN for classification of the neoplasm \cite{chen2018mri, 6165309}.

After a detailed analysis of the existing techniques, the following issues are found:
\begin{itemize}
\item In the literature, very few preprocessing techniques have been used like noise removal and contrast optimization. Further, many works in the literature skip this phase.
\item Many approaches in the literature used techniques that extracts texture and intensity-based features from MRI data to improve the classification performance, but these techniques are used assuming that the feature requirements are already known. Although these features provide improvement, however, there is a possibility of unknown features that might not have been considered.
\item PCA is a good feature selection algorithm as it tries to cover maximum variance among the features in a dataset. However, incorrect selection of the number of principal components can lead to information loss \cite{uyeda2015comparative}. Information loss in this domain can be costly and can affect the overall model performance.
\item Classification algorithms are optimized for generalized performance, but due to difference in the requirement of various domains, these generalized models may not be suitable like in the case of brain neoplasm MRI data classification \cite{pendharkar2003evolutionary}.
\item Some of the papers in the literature used the 3D-CNN model for MRI data. 3D-CNN is good for 3D image classification, but CNN, in general, suffers when the dataset size is not adequate\cite{gal2017deep}. Further, implementing an additional CNN is computationally costly and may suffer from the drawbacks of vanishing or exploding gradients. Identification of edge is a crucial task in brain neoplasm MRI classification but 3D CNN based method has a severe over-smoothing problem on edge boundaries\cite{DBLP:journals/corr/abs-1803-08669}.
\item Significance of model performance metrics are vastly different according to the domain of the work and accuracy may not be the ultimate evaluating metric for MRI data classification.
\end{itemize}


Preprocessing techniques in the literature consider noise removal, preserving texture information but does not consider edge detection, a technique considering edge is missing in the literature. The majority of the papers published in these areas have extracted mainly textural and wavelet-based features, but there is a possibility of other un-recognized features that may improve model performance. Further, these techniques require manual intervention and can be challenging when considering new data. Not many hybrid techniques are used or compared in this field of research. For classification, the majority of the papers have tried only to maximize accuracy assuming that false positives and false negatives are equally bad. However, in the medical domain, these two types of errors have very different costs associated. Concerning these limitations, two main objectives are drawn:
\begin{enumerate}
\item Propose a hybrid preprocessing stage to improve the area of interest in MRI Image.
\item Propose a CSVM model for harnessing the ability of CNN for feature extraction and SVM for classification.
\item Increasing model performances taking into consideration the different costs associated with false positives and false negatives.
\end{enumerate}

\section{Related Work}
\label{S:2}
In this section, a few important works related to brain neoplasm detection are briefly surveyed.
\vspace{-0.5  cm}
\subsection{Preprocessing techniques}
\label{S:2.1}
MRI data may contain noise and defects that can hamper model performance. To deal with such noisy data, various preprocessing techniques have been proposed in the literature.  

Amin and Mageed \cite{amin2012brain} proposed image preprocessing using image cutting, image enhancement, and size normalization. They used histogram equalization to increase the contrast range by increasing the dynamic range of grey levels. They showed improvement in the distinction of features in the MRI image. 

Gadpayleand and Mahajani \cite{mahajanidetection} used Gaussian filter to remove granular noise and resized the MRI data due to large size variations in the used dataset. They used manual cropping for some MRI data for convenience. 

George and Karnan \cite{george2012mri} used a modified Tracking Algorithm to remove the film artifacts and then applying the Histogram Equalization and Center Weighted Median (CWM) filter techniques separately to enhance the MRI images. 

\subsection{Feature optimization techniques}
\label{S:2.2}

MRI data contains a large number of features and using all features not necessarily increases the performance of a classification algorithm. Reducing the number of features in the dataset usually improves accuracy as well as training and testing time. An efficient way to select important features is PCA ()Principle Component Analysis). PCA technique can remove multicollinearity in the dataset, improves algorithm performance, and reduces over-fitting. Further, the texture is an important aspect of MRI data, and to extract the texture features, GLCM (Gray Level Co-occurrence Matrix) is widely used, especially in medical image analysis.

Amin and Mageed \cite{amin2012brain} proposed a neural network and segmentation base system to automatically detect the tumor in brain MRI images. In their approach, PCA is used for feature extraction. Othman et al. \cite{othman2011probabilistic} proposed a probabilistic neural network technique for brain tumor classification. In their technique, features are extracted using PCA. Daljit Singh et al. \cite{singh2012classification} proposed a hybrid technique for automatic classification of an MRI image by extracting features using PCA and GLCM. Joshi et al. \cite{joshi2010classification} proposed a brain tumor detection and classification system in MR images by first extracting the tumor portion from the brain image, then extracting the texture features of the detected tumor using GLCM. Gadpayleand and Mahajani \cite{mahajanidetection} proposed a brain tumor detection and classification system. The tumor is extracted using segmentation and then texture features are extracted using GLCM.

Kharrat \cite{kharrat2010hybrid} proposes a hybrid approach for the classification of brain tissues. The classification is based on a Genetic algorithm (GA) and support vector machine (SVM). Features considered in their work are wavelet-based texture features. They extracted features by spatial gray level dependence method (SGLDM). The extracted features were given as input to the SVM classifier.

\vspace{-0.25  cm}
\subsection{Classification techniques}
\label{S:2.3}
Many techniques are used in the literature for MRI brain neoplasm classification. The objective in these approaches is to detect the presence and/or severity of brain neoplasm.\par

Amin and Mageed \cite{amin2012brain} used Multi-Layer Perceptron (MLP) to classify the extracted features of MRI brain images. The average recognition rate is 88.2\% and the peak recognition rate is 96.7\%. Othman et al. \cite{othman2011probabilistic} performed brain tumor presence classification using Probabilistic Neural Network (PNN). Daljit Singh et al. \cite{singh2012classification} used a Support Vector Machine (SVM) classifier which classifies the image to either normal or abnormal. Joshi et al. \cite{joshi2010classification} detected the presence of brain tumor using the neuro-fuzzy classifier. Gadpayleand and Mahajani \cite{mahajanidetection} combined the BPNN and KNN classifiers to classify MRI brain images into the normal or abnormal brain. The accuracy is 70\% using the KNN classifier and 72.5\% by using the BPNN classifier. Kharrat \cite{kharrat2010hybrid} used the binary SVM classifier with RBF kernel to detect the presence of a brain tumor with an accuracy rate that varies from 94.44\% to 98.14\%. Jafari and Shafaghi \cite{jafari2012hybrid} proposed a hybrid approach for brain tumor detection in MR images based on Support Vector Machines(SVM). The accuracy of about 83.22\% is achieved and claimed their approach robust in detection.

\vspace{-0.25  cm}
\section{Proposed Methodology}
\label{S:3}

\begin{figure}[t]
  \centering
  \includegraphics[width=6.0 cm]{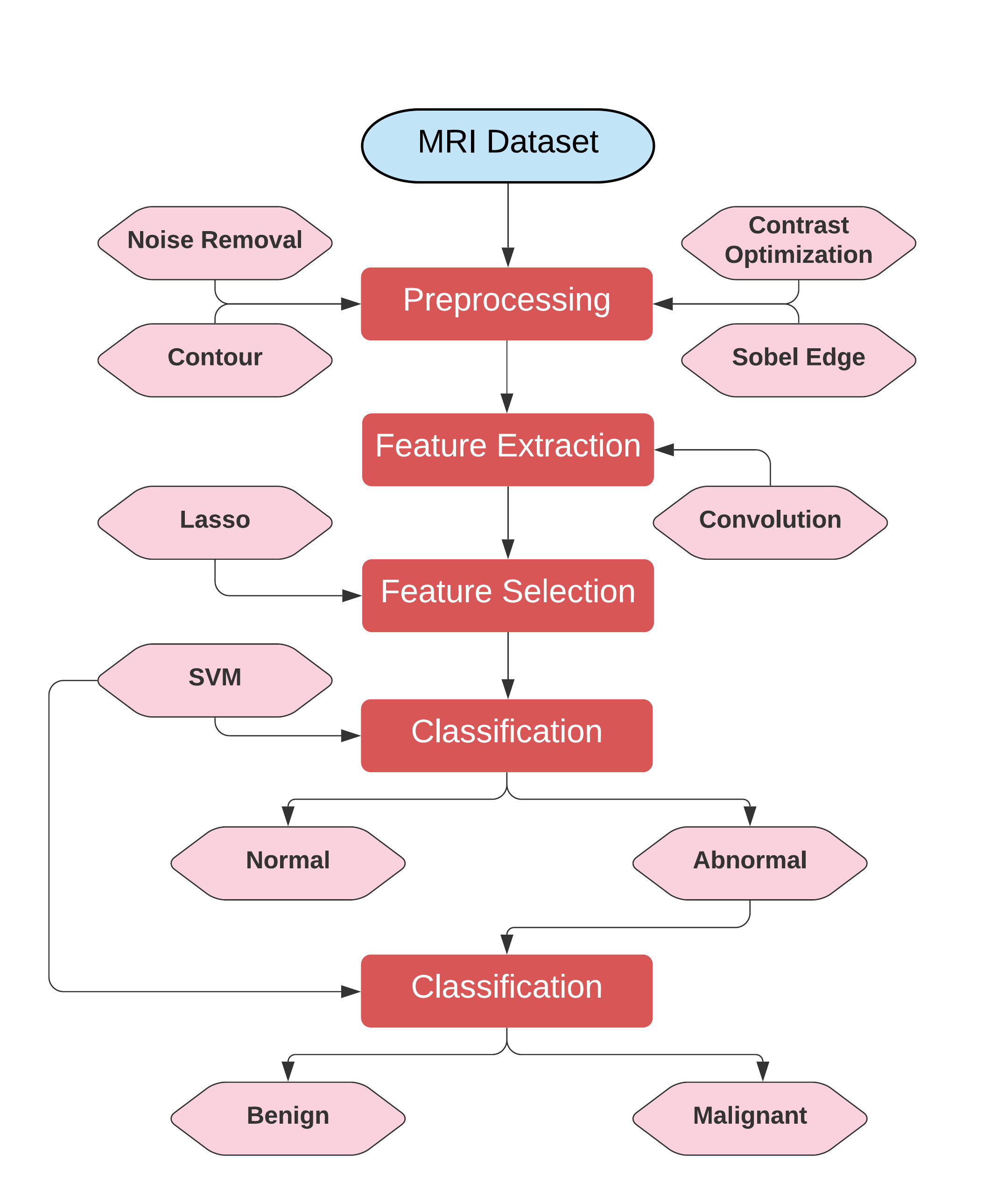}
  \caption{Overview of the proposed brain neoplasm detection method.}
  \label{fig:architecture}
\end{figure}

A computational intelligence approach is proposed to provide better and reliable classifications of brain neoplasms in MRI data. An overview of the proposed method is shown in Fig. \ref{fig:architecture}. The major computational steps in the proposed approach are image preprocessing, feature engineering, and classification. The various steps in the approach are stated below:
\vspace{-0.25 cm}
\begin{enumerate}
\item \textbf{\textit{Preprocessing}}: The preprocessing step consists of two sub-steps. First, the basic preprocessing techniques like resizing, noise removal, and contrast optimization are performed. Next, a preprocessing technique is proposed based on Sobel edge detection and contour to improve the area of interest in the MRI image.

\item \textbf{\textit{Feature engineering}}: The proposed feature engineering includes feature identification, feature extraction followed by feature optimization. We propose an automatic approach to feature identification and extraction. The convolution neural network model is built to generate important features automatically for the brain neoplasm MRI data. Following this LASSO regularization technique is used to remove unnecessary features and select the suitable and most dominating features.

\item \textbf{\textit{Classification}}: We propose a two-stage classification approach. Two classification models are used for the purpose. In the first stage of classification, whether an input MRI data contains a tumor or not is predicted. If an image contains a brain neoplasm, then the image is passed through the second stage of the classifier. In the second stage of the classification, the severity of the tumor is classified as benign or malignant. In both stages of the classification, the SVM model with modified cost optimization constraint is considered to minimize the false negative classification. \\
    
    In the following subsections, the above-mentioned three steps are discussed in details.
    
\end{enumerate}

\subsection{Data preprocessing}
\label{S:3.2}
There are two approaches to data preprocessing: regular data preprocessing and specific preprocessing. The regular preprocessing task is to deal with poor quality MR images. The improved quality MR image is used in the next level of the preprocessing step to specify a region of interest. The next level of preprocessing is proposed to improve the prediction accuracy.

\subsubsection{Basic preprocessing}
\label{S:3.2.1}
The basic preprocessing includes resizing, noise removal, and contrast enhancement.  The methodologies proposed for these activities are  discussed in the following section.

\begin{figure}[H]
  \centering
  \includegraphics[width=10cm]{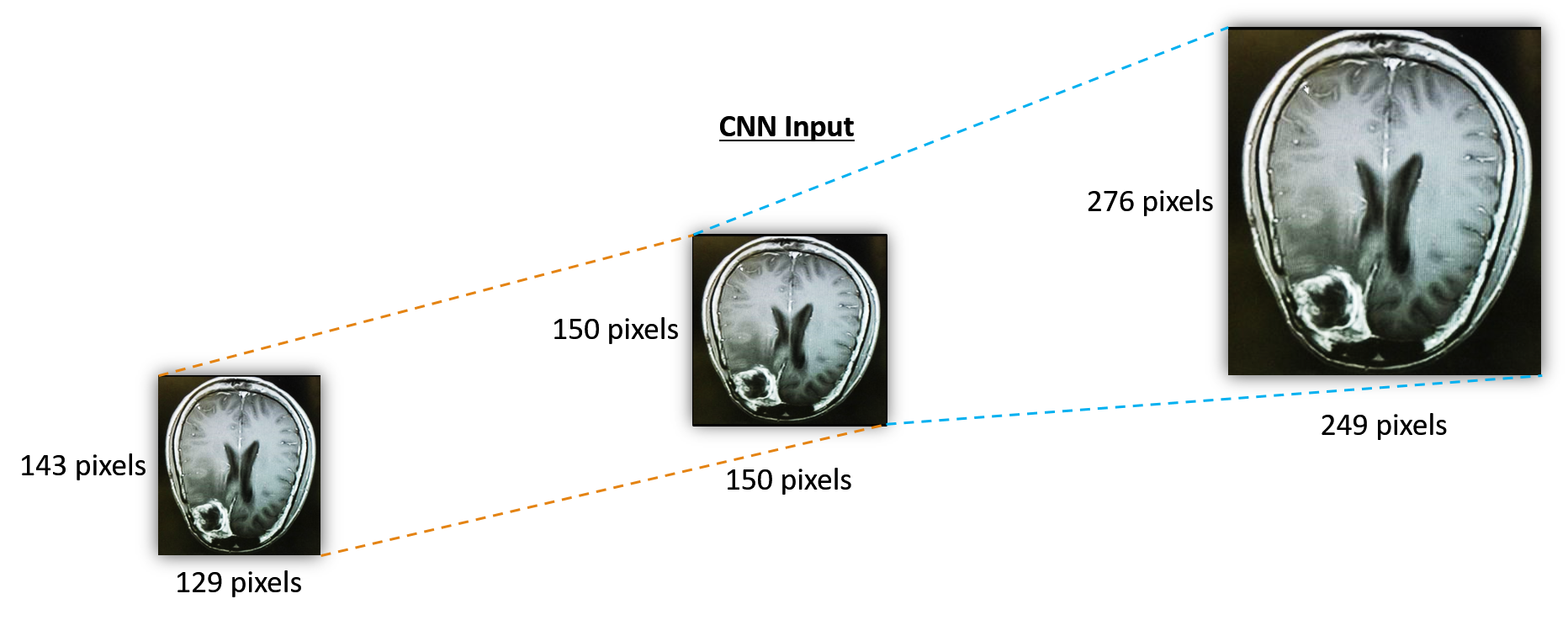}
  \caption{Example of resizing an MRI image.}
  \label{fig:res_pp}
\end{figure}

\textbf{\textit{Resizing:}} It is the process of changing the scale of the images so that all the
samples in the training and testing data sets are of the same resolution. This is required because, in the classification models, we must provide input of the same size. The images are resized to $150\times150$ pixels using the using Numpy squeeze function for each MRI data, as shown in Fig. \ref{fig:res_pp}.

\begin{figure}[H]
  \centering
  \includegraphics[width=8cm]{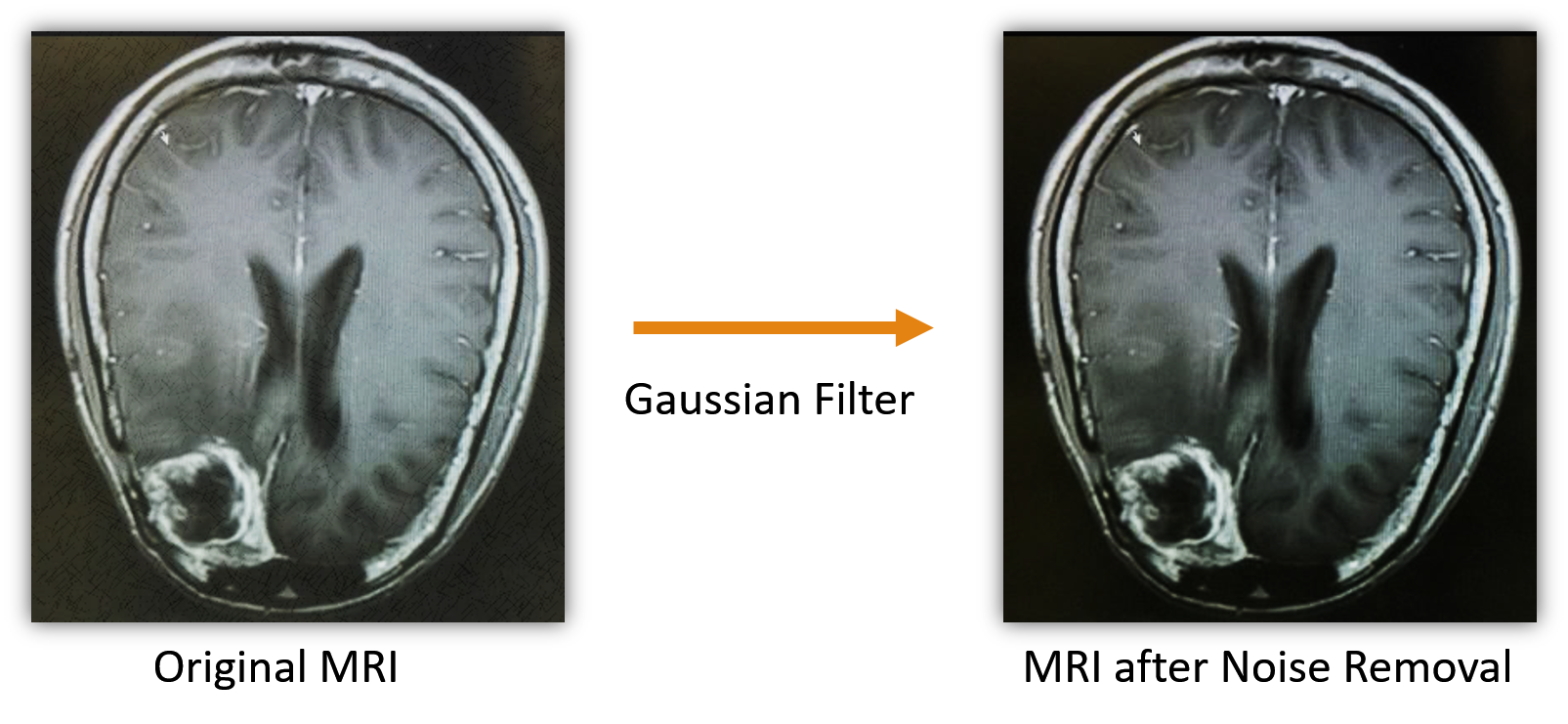}
  \caption{Example of noisy MR image and the image after noise removal.}
  \label{fig:noise_pp}
\end{figure}

\textbf{\textit{Noise removal:}} Gaussian filter is a well-known method and is used to smooth a noise by taking the average values surrounding a noise \cite{kumar2017noise}. In this experiment, the noise removal is performed using a Gaussian filter. Essentially, a Gaussian filter is a low pass filter that is non-uniform. Also, the kernel is symmetric in a Gaussian filter, which can reduce direction bias, if any. Application of a Gaussian filter requires convolution of 2D Gaussian distribution (see Eqn. \ref{eqn:2d_gaussian}) given an image input. It may be noted that 2D Gaussian distribution is the product of two 1D Gaussian functions (see Eqn. \ref{eqn:1d_gaussian}). The kernel coefficients are sampled from Eqn. \ref{eqn:2d_gaussian}. An example showing the noisy image and the same after the removal of noise is shown in Fig. \ref{fig:noise_pp}.

\begin{equation}
    G\left ( x \right ) = \frac{1}{\sqrt[]{2\pi \sigma ^{2}}}e^{-\frac{x^{2}}{2\sigma ^{2}}}
    \label{eqn:1d_gaussian}
\end{equation}
\begin{equation}
    G\left ( x,y \right )=\frac{1}{2\pi \sigma ^{2}}e^{-\frac{x^{2}+y^{2}}{2\sigma ^{2}}}
    \label{eqn:2d_gaussian}
\end{equation}

\textbf{\textit{Contrast enhancement:}} A Gaussian filter might hamper the image brightness and contrast. Hence, a contrast optimization procedure is required. The contrast of an image is the difference in the luminance of various sections in the image. An optimal contrast-enhanced image will be able to show every object present in the image. The need for contrast enhancement technique in medical field images \cite{kotkar2013review} are of utmost importance. Local and global transformation techniques \cite{kabir2010brightness} prove to be the best options in this regard as they tend to maintain the mean brightness and does not inject undesirable artifacts\cite{kotkar2013review}. Global Transformation Histogram Equalization is used for contrast enhancement in this work. The global transformation function formulated in Eqn. \ref{eqn:global_contrast} is used for contrast enhancement. Here, $T(g)$ being the global transformation function, where \textit{g} denote its intensity, $g_{min}$ and $g_{max}$ defines the lower and upper bounds of the histogram partitions; here, x as intensity value, h(x) denotes the histogram count.

\begin{equation}
    T\left ( g \right ) = g_{min} + \left ( g_{max} - g_{min} \right ) \left ( \frac{\sum_{x = g_{min}}^{g} h\left ( x \right ) }{\sum_{x = g_{min}}^{g_{max}} h\left ( x \right )} \right )
    \label{eqn:global_contrast}
\end{equation}

\begin{figure}[ht]
  \centering
  \includegraphics[width=8cm]{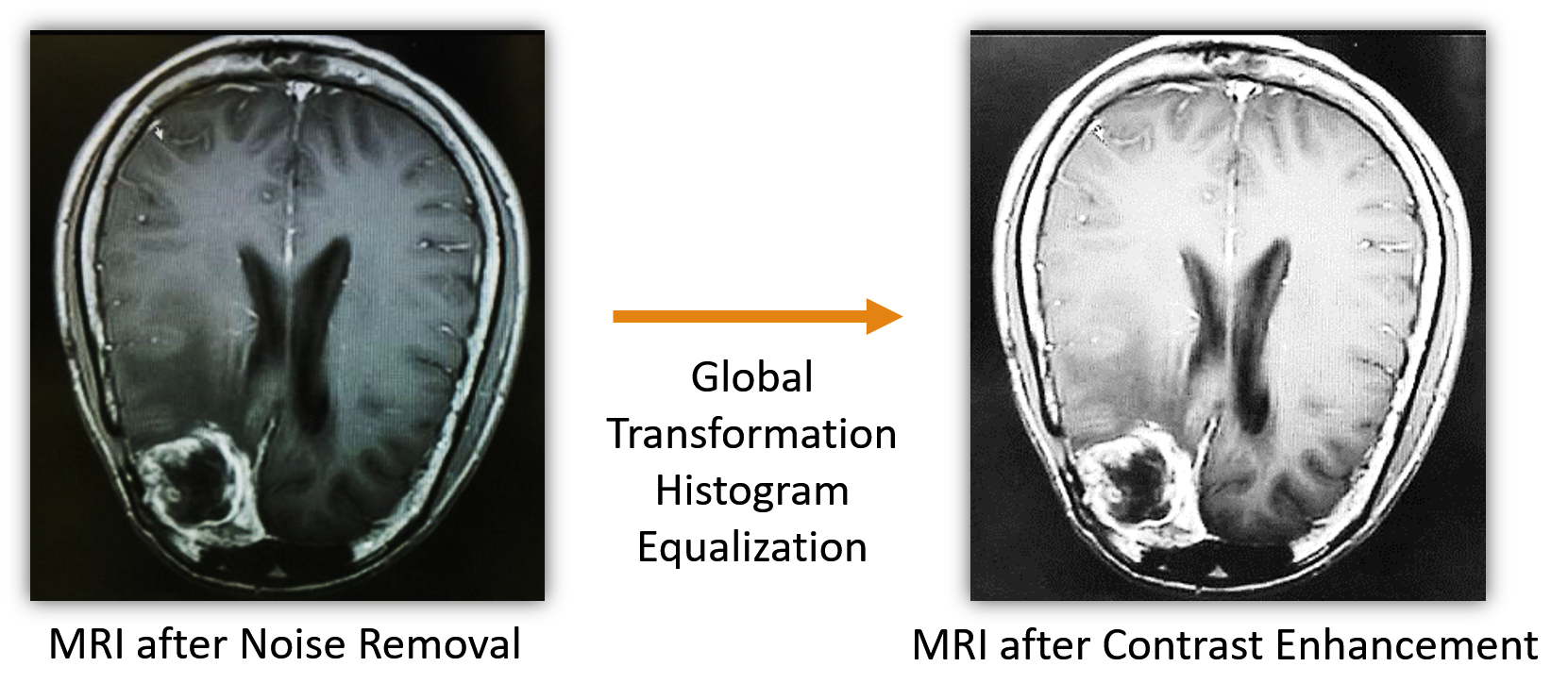}
  \caption{Example of contrast enhancement in MRI image}
  \label{fig:cont_pp}
\end{figure}

An example showing the contrast enhancement using Global Transformation Histogram Equalization is shown in Fig. \ref{fig:cont_pp}.

\subsubsection{Special preprocessing}
\label{S:3.2.2}
\begin{figure}[H]
  \centering
  \includegraphics[width=9cm]{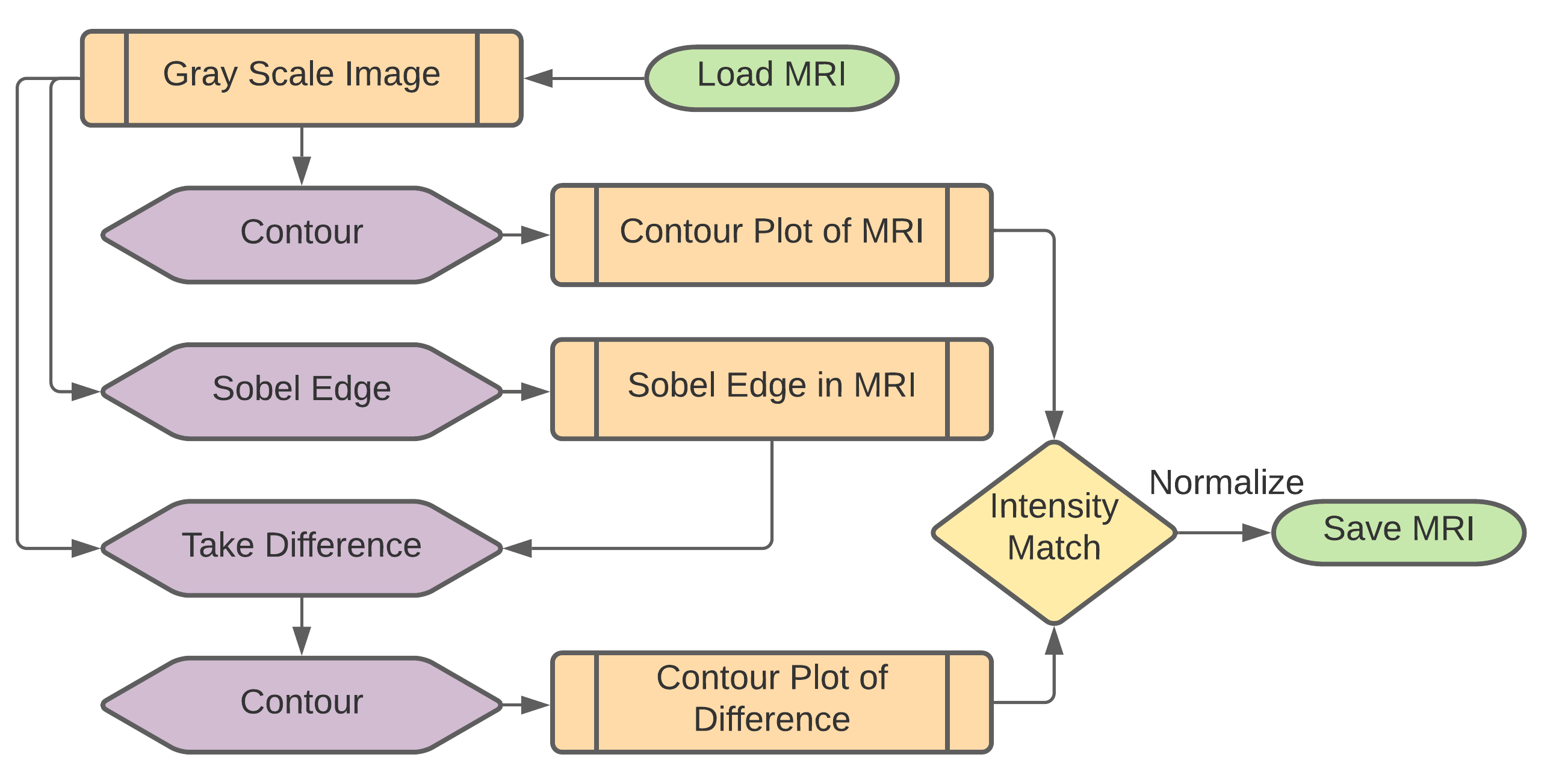}
  \caption{The proposed Sobel edge detection based preprocessing.}
  \label{fig:prepro_model}
\end{figure}

Apart from the regular preprocessing techniques, an extra filtering mechanism is proposed. The proposed technique is shown in Fig. \ref{fig:prepro_model}. The proposed technique starts with loading an MRI data which is already preprocessed using the basic techniques described in Section \ref{S:3.2.1}. Then a contour plotting operation is performed on the MRI data. In the next step, the Sobel edge detection technique is used on the grayscale image to generate edges. Further, the Sobel edge MRI data matrix is subtracted from the grayscale image data matrix. This image is matched with the contour plot in the first step to preserve brightness. Then the final image is stored in a grayscale format. This preprocessing helped in reducing the already present artifacts and proved to improved the features of the region of interest in MRI data. The contour plot, Sobel edge detector, and the different functions are further described in detail below.

\begin{figure}[b]
	\begin{subfigure} {1.0\textwidth}
  \centering
  \includegraphics[width=5.5cm]{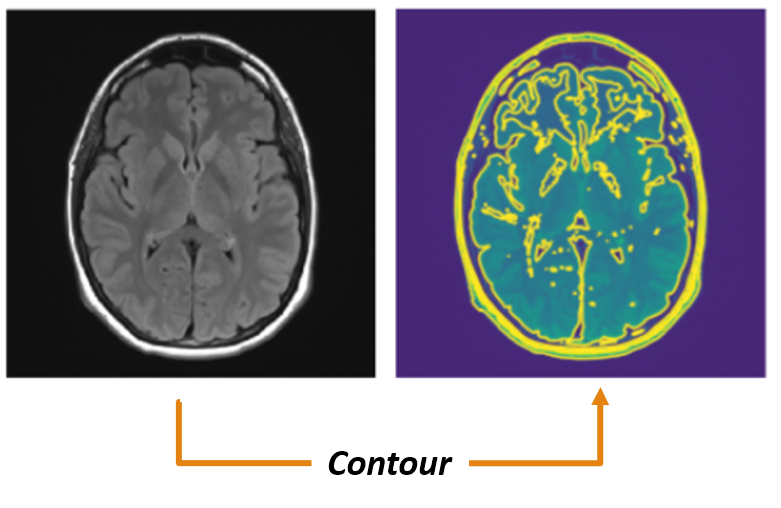}
  \caption{Contour plot for gray scale MRI image.}
  \label{fig:ex_contour}
\end{subfigure}

\begin{subfigure} {1.0\textwidth}
\centering
\includegraphics[width=5.5cm]{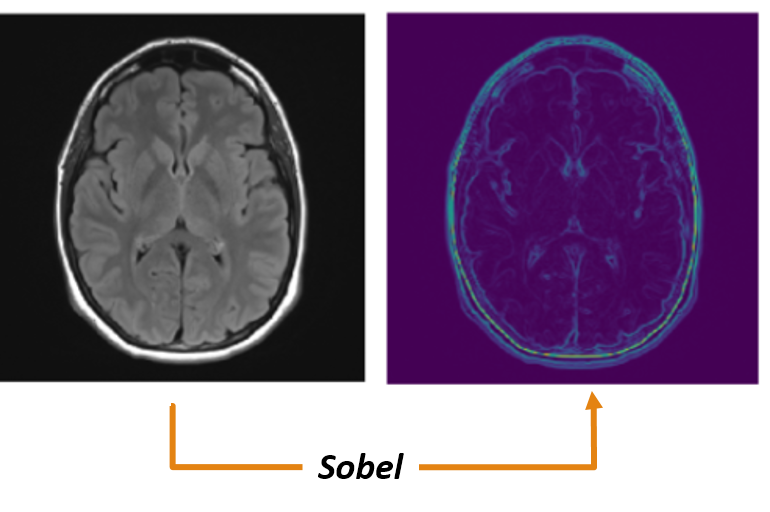}
\caption{Sobel edge detection for gray scale MRI image.}
\label{fig:ex_sob}
\end{subfigure}

\begin{subfigure} {1.0\textwidth}
		\centering
		\includegraphics[width=5.5cm]{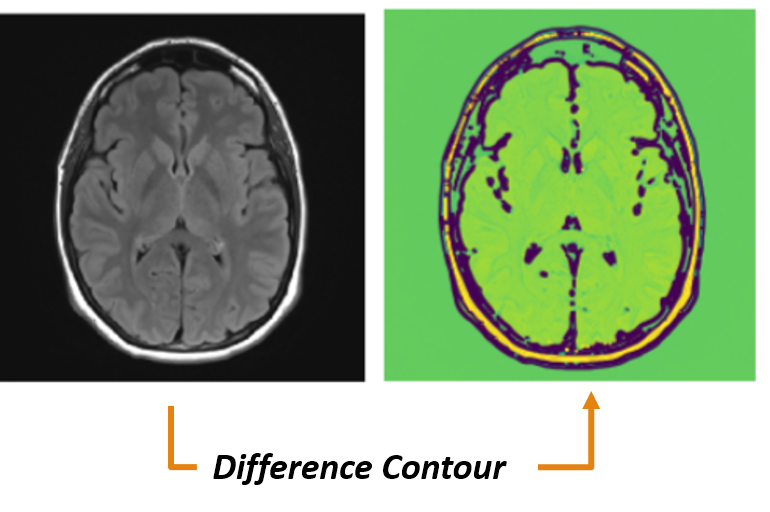}
		\caption{Difference contour for gray scale MRI image}
		\label{fig:ex_dif}
\end{subfigure}
\caption{An example  of special preprocessing technique.}
\end{figure}

\textbf{Contour calibration:}
Regarding MRI images, it is desired to have a higher contrast in the area of interest and lower on the rest. The images that are analyzed till now are in gray scale, hence the different variations in textures are not visible. To know the depth or height of a 3D plane contour lines can be used. It is a function of two variables in a curve along which the function has a constant value so that the curve joins points of equal value. A gray-scale image can also be seen as a 3D representation of the values ranging from 0 to 255 for each pixel of which contour can be plotted\cite{kim2014relationship}. In this experiment, the MRI has a size of 150 $\times$150 and the contour line can be used to define various colors to the images with varying gray scale. An example showing the contour plot for an MRI sample is shown in Fig. \ref{fig:ex_contour}.

\textbf{Sobel edge detection:}
Sobel edge detection technique is highly efficient in the detection of edges in images if the image is free of granular noises. Modified Sobel edge detection techniques\cite{gao2010improved} proved to perform superior in certain conditions. In this experiment, after the basic preprocessing of the MRI, granular noise is certainly removed in the Gaussian filtering process. Hence, we can proceed with the basic Sobel edge detection technique \cite{vijayarani2013performance, vincent2009descriptive}for detection of edges present in the MRI images. An example showing the Sobel edge detection for an MRI sample is shown in Figure-\ref{fig:ex_sob}.


\textbf{Differencing:}
In this phase, the matrices of grayscale MRI image is subtracted from the Sobel edge MRI image are on a pixel by pixel basis. If different contour intensity is present in original contour and Sobel edge contour then take the higher value of intensity. This phase reduced the less oriented regions and gave proper intensity uplifting in the regions of interest. An example showing the contour matching operation over the difference image for an MRI sample is shown in Fig. \ref{fig:ex_dif}.


\subsection{Feature engineering}
\label{S:3.3}
Due to the lack of a huge amount of MRI data in a single place, it is very difficult to analyze and classify images related to brain neoplasm. The machine learning algorithms are optimized by using feature extraction \cite{guyon2008feature} techniques to work with even a very low amount of data and still performing quite well. Automatic feature extraction for brain neoplasm MRI data is performed using a convolution feature extractor. Convolutional neural networks are comprised of two parts, the first section performs convolution using various kernels to extract features and then classification is performed by dense layers. In a hybrid CSVM model, the first phase of a normal CNN is used for feature extraction and the features are then fed to SVM for further classification. Feature extraction using convolution neural networks have been performed in various fields of research due to its flexible and reliable results \cite{simard1999boxlets, hammad2019novel}. CNN model-based feature extractors also prove to be superior in extracting textures from images\cite{dewaele1988texture, zhang2000all}, which are a very important aspect in MRI classifications as well. The convolution feature extractor model for detecting the presence and severity of neoplasms in the MRI dataset is discussed below.

\begin{table}[b]
	\centering
	\resizebox{9cm}{!}{%
		\begin{tabular}{|ccc|}
			\hline
			\multicolumn{3}{|l|}{\textit{Model: sequential}} \\ \hline
			\multicolumn{1}{|c|}{\textbf{Layer Name\_Type} }& \multicolumn{1}{c|}{ \textbf{Dimension} } & \textbf{Parameters} \\ \hline
			\multicolumn{1}{|l|}{Block-1\_Conv-1 (Conv2D)} & \multicolumn{1}{c|}{ $(None, 148, 148, 32)$ } & 320 \\ \hline
			\multicolumn{1}{|l|}{Block-1\_MP-1 (MaxPooling2D)} & \multicolumn{1}{c|}{$(None, 74, 74, 32)$} & 0 \\ \hline
			\multicolumn{1}{|l|}{Block-2\_Conv-1 (Conv2D)} & \multicolumn{1}{c|}{$(None, 72, 72, 64)$} & 18496 \\ \hline
			\multicolumn{1}{|l|}{Block-2\_Conv-2 (Conv2D)} & \multicolumn{1}{c|}{$(None, 70, 70, 64)$} & 36928 \\ \hline
			\multicolumn{1}{|l|}{Block-2\_MP-1 (MaxPooling2D)} & \multicolumn{1}{c|}{$(None, 35, 35, 64)$} & 0 \\ \hline
			\multicolumn{1}{|l|}{Block-3\_Conv-1 (Conv2D)} & \multicolumn{1}{c|}{$(None, 33, 33, 96)$} & 55392 \\ \hline
			\multicolumn{1}{|l|}{Block-3\_MP-1 (MaxPooling2D)}     & \multicolumn{1}{c|}{$(None, 16, 16, 96)$} & 0\\ \hline
			\multicolumn{1}{|l|}{Block-4\_Conv-1 (Conv2D)} & \multicolumn{1}{c|}{$(None, 14, 14, 96)$} & 83040 \\ \hline
			\multicolumn{1}{|l|}{Block-4\_MP-1 (MaxPooling2D)}     & \multicolumn{1}{c|}{($None, 7, 7, 96)$} & 0\\ \hline
			\multicolumn{1}{|l|}{Block-5\_Conv-1 (Conv2D)} & \multicolumn{1}{c|}{$(None, 5, 5, 64)$ } & 55360 \\ \hline
			\multicolumn{1}{|l|}{Block-5\_MP-1 (MaxPooling2D)}     & \multicolumn{1}{c|}{ $(None, 2, 2, 64)$} &  0\\ \hline
			\multicolumn{1}{|l}{Total params: 249,536} & \multicolumn{1}{l}{}  & \multicolumn{1}{l|}{} \\
			\multicolumn{1}{|l}{Trainable params: 249,536} & \multicolumn{1}{l}{}  & \multicolumn{1}{l|}{} \\
			\multicolumn{1}{|l}{Non$-$trainable params: 0} & \multicolumn{1}{l}{}  & \multicolumn{1}{l|}{} \\ \hline
		\end{tabular}%
	}
	\caption{\label{tab:model_fex}Layer details of convolution model for feature extraction.}
\end{table}

\begin{figure}[hb]
  \centering
  \includegraphics[width=9cm]{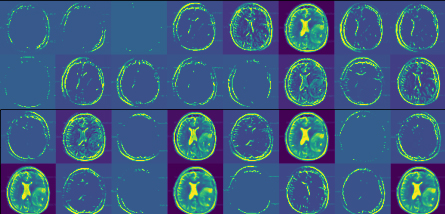}
  \caption{Features extracted from a single slice in \textit{Block-1\_Conv-1} layer.}
  \label{fig:features_one}
\end{figure}

Training with the MRI dataset, which is made up of 150$\times$150 gray-scale images, that is, each image is of shape (150,150,1) (Gray Scale = 1 channels). The sequential convolution feature extraction model is provided in Table \ref{tab:model_fex} containing layer names, dimension, and parameters. Further, note that the CNN used in the experiment is also having the same architecture apart from the SVM classifier, which is replaced using a fully connected neural network. An example of the features extracted using a particular convolution layer on a particular slice is given as an example in Fig. \ref{fig:features_one}. The generated feature vector will be used for the feature selection phase for reducing unnecessary features. LASSO regularization is used to select the optimal set of features and remove the features that may hamper the model performance. Convolutional feature extractor provides 2,49,536 features based on the total input data which has been significantly reduced to 5,240 features after LASSO feature selection is performed.

\vspace{-0.8cm}
\subsection{MRI data classification}
\label{S:3.5}
\begin{figure}
  \centering
  \includegraphics[width=9cm]{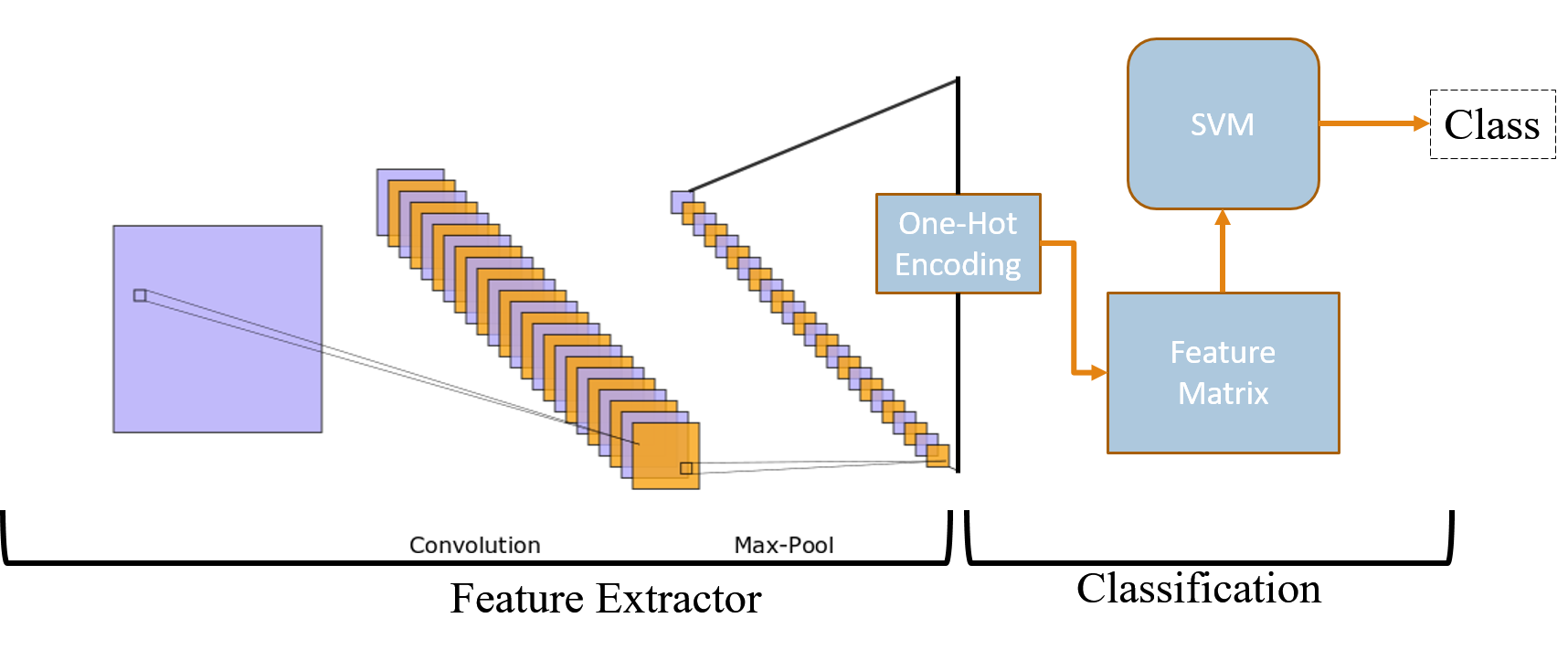}
  \caption{CSVM model for automatic feature extraction and classification.}
  \label{fig:csvm_model}
\end{figure}

To understand the presence of a brain tumor or to classify the severity to either benign or malignant is a binary classification problem. As per SVM (Support Vector Machine) is concerned, it is a highly accepted binary classifier for linear data and non-linear data(using special kernels). SVM uses the concept of finding the best hyperplane that can separate the data perfectly into two classes. SVM is a popular choice for classification irrespective of the application domains, which varies from text\cite{nilanjan17} to medical image classification\cite{machhale2015mri, singh2012classification}. Further, SVM tends to perform well even on the small datasets \cite{huang2017svm} but deep learning requires a huge amount of data\cite{gal2017deep}. The proposed approach (CSVM) using CNN feature extraction and SVM classification is shown in Fig. \ref{fig:csvm_model}.

SVM optimizes the hyperplane for the best classification of the classes without considering the various cost associated with false positives and false negatives. In general, this is not a problem as for many tasks there is no significant cost imbalance for classes. But for a medical domain such as the classification of brain MRI data for detection of neoplasm, the false negatives have a higher cost associated when compared with false positives. To deal with this issue of generalized SVM, the optimization function of the SVM needs to be modified.
The sample space is
   \{$x_i,y_i$\},$\mathrm{\textrm{i}}$ = 1,2, . . . , \textit{m}.
 where, y${}_{\mathrm{\textrm{i}}}$ 
  x${}_{\mathrm{\textrm{i}}}$$\mathrm{\in }$ $\mathrm{\{}$F${}_{1}$, F${}_{2}${\dots} F${}_{n}$$\mathrm{\}}$. The sample space has been separated with a hyperplane given by 
 g($\mathrm{\textrm{x}}$) = $w^Tx_i+b$ = 0
 where, \textit{w} is a \textit{n}-dimensional coefficient vector that is normal to the hyperplane and \textit{b} (bias) is the offset.

\textbf{\textit{Criteria for classification:}} Considering positive class indicates the presence of neoplasm for the first classifier model and indicates malignant for the severity classification model. The negative class on the other hand indicates an absence of neoplasm for the first model and benign for the severity classification of the neoplasm. Let \textbf{$\boldsymbol{\mathrm{\textrm{C}}}_{s}$} denote the unequal misclassification costs associated with the false negative class and \textbf{\textit{x${}_{i}$}} is not perfectly separable with the hyperplane, then some samples are allowed to be at a distance ${\mathrm{\xiup }}_i$ from their correct margin boundary. The primal problem of SVM has been modified into solving the following optimization task:

 \begin{center}
$f\left(x_i,y_i\right)=min \frac{1}{2}{\left\|w\right\|}^2+C_s\sum_{i|y_i}{{\mathrm{\xiup }}_i}$

\[subject~to,~~y_i\left[\left(w^Tx_i+b\right)\right]~\ge 1~-{\mathrm{\xiup }}_i\] 
\[where,~~i=1,2,~\dots ,~m~~~~~and~~~{\mathrm{\xiup }}_i\ge 0\] 
\end{center}
\begin{equation}
    \label{eq:opt}
\end{equation}

Considering \textit{C(penalty term)} which controls the strength of penalty for $\mathrm{\xiup }$. $C_s i|y_{i=+1}$ = C and $C_s i|y_{i=-1}$ = $C^r$. Where, $C^r=~\frac{C_s}{C}$ provided $C_s>C$ implying, $C^r>0$. Since, $f\left(x_i,y_i\right)$ is a constraint optimization problem, $\mathrm{therefore,}$ converting this to un-constraint optimization problem is done using Lagrangian Multiplier with $\boldsymbol{\alpha }~,~~\boldsymbol{\beta }$ as the Lagrange multipliers where $\boldsymbol{\alpha }\boldsymbol{\ge }\boldsymbol{0}~and~\boldsymbol{\beta }\boldsymbol{\ge }\boldsymbol{0}$.

\begin{center}
\[~\mathrm{L(}~\mathrm{w,}~\mathrm{b,}~C_i,~{\mathrm{\xiup }}_i\mathrm{)}= \]
= $\frac{1}{2}{\left\|w\right\|}^2$ + $\sum{C_i{\mathrm{\xiup }}_i}$ + $\sum{{\alpha }_i}$ - $\sum{{\alpha }_i{\mathrm{\xiup }}_i}~-~\sum{{\alpha }_iy_i(w^Tx_i)}$ $-$ $\sum{{\alpha }_iy_ib}$ $-$ $\sum{{\beta }_i}{\mathrm{\xiup }}_i$

\[C_i = \left\{ \begin{array}{c}
C~~~~~~~~~~~~~~~y_i\in +1 \\ 
C^r*C~~~~~y_i\in -1 \end{array}
\right.\] 
\end{center}
\begin{equation}
    \label{eq:optL1}
\end{equation}

Optimization of the above Lagrangian \eqref{eq:optL1} involves taking derivatives of the Lagrangian with respect to w, b, $C_i$ and ${\mathrm{\xiup }}_i$ and set the derivatives to zero to get the below equation:
\begin{center}
 $\mathrm{L}~\mathrm{(}~\mathrm{w,}~\mathrm{b,}~C_i,~{\mathrm{\xiup }}_i\mathrm{)}=\sum_i{{\alpha }_i}-\sum_i{{\sum_j{{\alpha }_i}\alpha }_jy_iy_jK\left(x^{(i)},x^{(j)}\right)}$

\[subject~to~\left\{ \begin{array}{c}
{\sum_i{{\alpha }_i}y}_i=0 \\ 
0~\le {\alpha }_i\le C_i \end{array}
\right.\] 
\end{center}
\begin{equation}
    \label{eq:optL2}
\end{equation}

Considering MRI brain neoplasm classification , $\mathrm{\textrm{f}}:\mathrm{\textrm{X}}\mathrm{\to}\mathrm{\textrm{Y}}$ is a nonlinear classification problem, SVM kernels needs to be used. Basically, a kernel function maps the nonlinear data to an n-dimensional space where the data is linearly separable by a hyperplane. SVM has many kernels to perform the optimizations like linear, polynomial, and radial basis function(RBF). The selection of a kernel proves to have significant variations in the performance of the models\cite{scholkopf2002learning}. RBF kernel maps the nonlinear vectors to a high dimension space where the best hyper pane is drawn and it proves to be the most preferred choice for image classification problems \cite{kharrat2010hybrid, kuo2013kernel,cao2008approximate, scholkopf2002learning}. Therefore, the Radial Basis Function (RBF) kernel function is used for the experiment.
\begin{center}
    $K\left(x_i,x_j\right)=exp\left(-g{\left\|x_i-x_j\right\|}^2\right)$

 where,
    ${\left\|x_i-x_j\right\|}^2$ is two-norm distance and,\\
    $\mathrm{g}$ is the kernel function parameter and $g=~\frac{1}{{2\sigma }^2}$

\end{center}
\begin{equation}
    \label{eq:rbf}
\end{equation}

 For an unknown feature vector `$x_j$', the classification decision with RBF Kernel \eqref{eq:rbf} is represented as :
\begin{center}
  D(z) = $w\bullet z+b$ =\textit{  }$\sum^m_{i=1}{{\alpha }_iy_i}$\textit{K(}$x_i\bullet x_j$\textit{) + b}
\end{center}
\begin{equation}
    \label{eq:dz}
\end{equation}

The parameters, \textbf{\textit{kernel function parameter }}$'g'$ and \textbf{\textit{penalty}} (C) needs to be tuned for the model to perform optimal throughout the dataset. For tuning the SVM model parameters, the K-fold cross-validation technique with the value of `\textit{k'} as 10 is used. Cross-validation provided the desired value of parameters C and $'g'$ to be 13 and 2 for detecting the presence of neoplasm and 9 and 3 for severity classification respectively, which are used for the rest the experiment.

\vspace{-0.5cm}
\section{Experiments and Experimental Results}
\label{S:4}
To establish the efficacy of the proposed approach, several experiments have been carried out. In this section, the objectives of the experiments, experimental environment, data sets used in the experiments, and results observed are discussed.

\subsection{Objectives of the experiments}
\label{S:4.1}
\vspace{-0.3cm}
Following are the objectives of our experiments:
\begin{enumerate}
    \item \textbf{Effectiveness of the proposed preprocessing:} Visual inspection and model training evaluation of proposed preprocessing model.
    \item \textbf{Effectiveness of the proposed cost optimization:} Performance difference of CSVM with and without cost-optimization.
    \item \textbf{Classification accuracy:} Performance evaluation of CSVM, CNN, SVM and Random Forest on the MRI data considering cost imbalance of classes.
    \item \textbf{Comparison with the existing approaches:} Performance comparison of CSVM with the existing works.
\end{enumerate}

\vspace{-0.5cm}
\subsection{Experimental environment}
\label{S:4.2}
\vspace{-0.3cm}
The specification of the system used is as follows:
O.S. - Windows 10 Professional, 
CPU - AMD® Ryzen™ 7-3700X Processor, 
RAM - 32GB DDR4, 
GPU - NVIDIA GeForce® GTX 1080 Ti. The windows version of the Python-64Bit\cite{10.5555/1593511} with editor IPython notebook \cite{PER-GRA:2007} throughout the experiment. Important modules used in the experiment include tensorflow \cite{tensorflow2015-whitepaper}, packages from NumPy \cite{oliphant2006guide}, SciPy \cite{2020SciPy-NMeth} and Matplotlib \cite{Hunter:2007}.

\vspace{-0.5cm}
\subsection{Dataset description}
\label{S:4.3}
\begin{figure}[H]
  \centering
  \includegraphics[width=6cm]{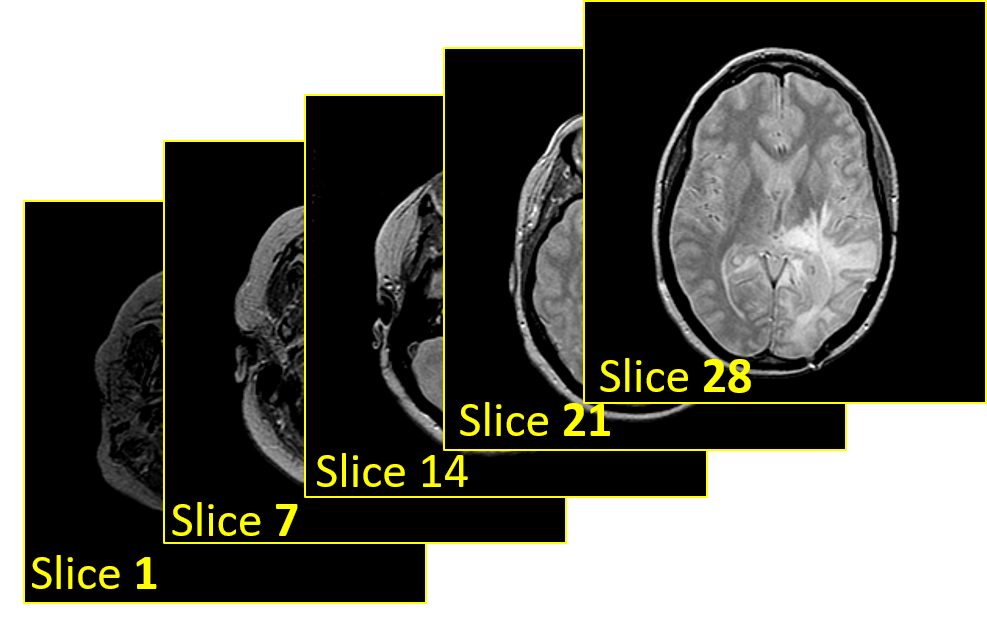}
  \caption{Example of stacked brain MRI slices}
  \label{fig:ex_data}
\end{figure}

\vspace{-0.3cm}
The objective of our work is to classify MRI images to predict the presence of neoplasms and their severity. In this experiment, an MRI dataset is collected from the online repository of The Whole Brain Atlas\cite{johnson_becker}. This website is maintained by Harvard Medical School and has proved to be very authentic in terms of medical imaging data\cite{summers2003harvard}.
T2 weighted images are used in the experiment as shown in Fig. \ref{fig:ex_data}. The types of the MRI for malignant brain neoplasms in the dataset are glioma and sarcoma. And the benign brain neoplasms are metastatic bronchogenic carcinoma(M.B.C.), metastatic adenocarcinoma(M.A.) and meningioma. The shape of an MRI image is \textit{h} $\times$ \textit{w} $\times$ \textit{n}, where \textit{n} is the number of slices in each image, \textit{h} is height and \textit{w} is width, where \textit{h} and \textit{w} both equals to 256.

\subsection{Results observed:}
\label{S:4.4}
Vis-a-vis the objectives of the experiments, the results which we have observed are presented in the following subsections. 
\subsubsection{Effectiveness of the proposed preprocessing}
\label{S:4.4.1}

\begin{figure}[hb]
     \centering
     \begin{subfigure}[b]{0.24\textwidth}
         \centering
         \includegraphics[width=\textwidth]{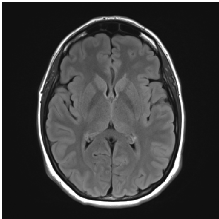}
         \caption{Normal GS}
         \label{fig:ngs}
     \end{subfigure}
     \hfill
     \begin{subfigure}[b]{0.24\textwidth}
         \centering
         \includegraphics[width=\textwidth]{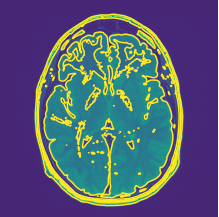}
         \caption{Normal CT}
         \label{fig:nct}
     \end{subfigure}
     \hfill
     \begin{subfigure}[b]{0.24\textwidth}
         \centering
         \includegraphics[width=\textwidth]{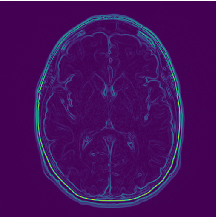}
         \caption{Normal SE}
         \label{fig:nse}
     \end{subfigure}
     \hfill
     \begin{subfigure}[b]{0.24\textwidth}
         \centering
         \includegraphics[width=\textwidth]{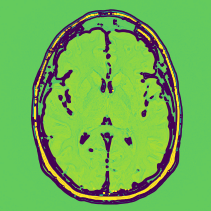}
         \caption{Normal SD}
         \label{fig:nsd}
     \end{subfigure}
        \caption{Proposed preprocessing applied over normal MRI images.}
        \label{fig:pro_nor}
\end{figure}

\begin{figure}[ht]
     \centering
     \begin{subfigure}[b]{0.24\textwidth}
         \centering
         \includegraphics[width=\textwidth]{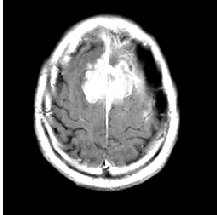}
         \caption{Benign GS}
         \label{fig:bgs}
     \end{subfigure}
     \hfill
     \begin{subfigure}[b]{0.24\textwidth}
         \centering
         \includegraphics[width=\textwidth]{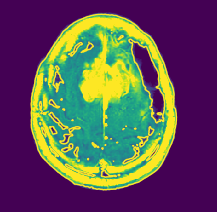}
         \caption{Benign CT}
         \label{fig:bct}
     \end{subfigure}
     \hfill
     \begin{subfigure}[b]{0.24\textwidth}
         \centering
         \includegraphics[width=\textwidth]{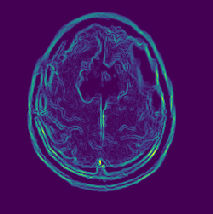}
         \caption{Benign SE}
         \label{fig:bse}
     \end{subfigure}
     \hfill
     \begin{subfigure}[b]{0.24\textwidth}
         \centering
         \includegraphics[width=\textwidth]{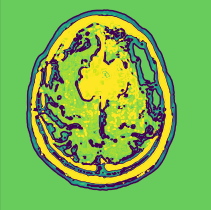}
         \caption{Benign SD}
         \label{fig:bsd}
     \end{subfigure}
        \caption{Proposed preprocessing applied over benign MRI images.}
        \label{fig:pro_ben}
\end{figure}

\begin{figure}[ht]
     \centering
     \begin{subfigure}[b]{0.24\textwidth}
         \centering
         \includegraphics[width=\textwidth]{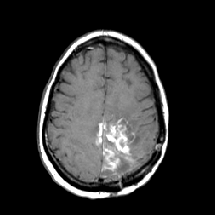}
         \caption{Malignant GS}
         \label{fig:mgs}
     \end{subfigure}
     \hfill
     \begin{subfigure}[b]{0.24\textwidth}
         \centering
         \includegraphics[width=\textwidth]{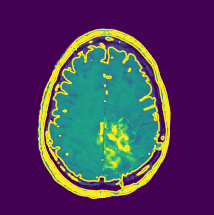}
         \caption{Malignant CT}
         \label{fig:mct}
     \end{subfigure}
     \hfill
     \begin{subfigure}[b]{0.24\textwidth}
         \centering
         \includegraphics[width=\textwidth]{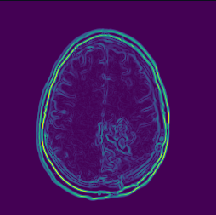}
         \caption{Malignant SE}
         \label{fig:mse}
     \end{subfigure}
     \hfill
     \begin{subfigure}[b]{0.24\textwidth}
         \centering
         \includegraphics[width=\textwidth]{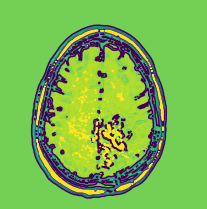}
         \caption{Malignant SD}
         \label{fig:msd}
     \end{subfigure}
        \caption{Proposed preprocessing applied over malignant MRI images.}
        \label{fig:pro_mal}
\end{figure}

\textbf{\textit{Visual inspection:}} An overview of the preprocessing for normal MRI can be seen in Fig. \ref{fig:pro_nor}. The preprocessing for benign MRI can be seen in the Fig. \ref{fig:pro_ben} and malignant MRI can be seen in the Fig. \ref{fig:pro_mal}.\\
To have a good demonstration of the preprocessing, three MRI images are considered, one is normal which is shown in Fig. \ref{fig:ngs}, that is, without any neoplasm, the second one with benign neoplasm which is shown in Fig. \ref{fig:bgs} and the third one with malignant neoplasm which is shown in Fig. \ref{fig:mgs}.The contour is plotted for three different MRI images, each for normal (see Fig.\ref{fig:nct}) , benign (see Fig.\ref{fig:bct}), and malignant (see Fig. \ref{fig:mct}).
Similar to contour, Sobel egdes are shown for the three MRI images: normal ()in Fig. \ref{fig:nse}), benign (in Fig.\ref{fig:bse}), and malignant (in Fig. \ref{fig:mse}).
The final result after taking difference is shown for the three MRI images, normal (Fig. \ref{fig:nsd}), benign (Fig. \ref{fig:bsd}), and malignant (Fig. \ref{fig:msd}).\\

To compare the preprocessed MRI images, visual inspection of image outline, the colour of the various regions needs to be explored. The green colour represents low intensity, yellow signifies higher intensity and brown lines indicate higher variation in the intensities over that region. Figure-\ref{fig:pro_nor} shows a transition of the various stages of the proposed preprocessing for a normal brain. The contour of the normal brain produces many yellow lines as seen in Figure-\ref{fig:nct}. But when compared with the difference contour, the regions are equally coloured without any yellow lines. Further, no brown edge is seen in the final image as seen in Figure-\ref{fig:nsd}. This means the proposed preprocessing can normalize the lower ranges of intensity difference (yellow lines).\\

\vspace{-0.3cm}

Figure \ref{fig:pro_ben} shows a transition of the various stages of the proposed preprocessing for a benign brain neoplasm. The contour of the benign brain produces many yellow regions as shown in Fig. \ref{fig:bct}. But when compared with the difference contour, the less interesting regions are colored green with a brown border. Further, a brown edge with yellow is is seen in the final image where the benign neoplasm is present as seen in Fig. \ref{fig:bsd}. This means the proposed preprocessing can intensify the regions with benign neoplasm than the other regions.

\vspace{-0.3cm}
Figure-\ref{fig:pro_mal} shows a transition of the various stages of the proposed preprocessing for a malignant brain neoplasm. The contour of the malignant brain produces a centralized yellow region as seen in Fig. \ref{fig:mct}. But when compared with the difference contour, the less interesting regions are colored green with a brown border as previously seen with benign MRI. Further, a brown edge with yellow is is seen in the final image where the malignant neoplasm is present as seen in Fig. \ref{fig:msd}. The proposed preprocessing can outline the portion with the brain tumor effectively, even if the tumor is discreetly spaced. Further, many preprocessing techniques tend to over soften the image during the noise removal process to get better accuracy, but this changes the original image features. This proves to be beneficial when detecting the presence of neoplasm but during the classification of the severity, this process fails to get good results. But in the case of the proposed preprocessing, the final output MRI image is free of over softness and hence the features are preserved better.

\begin{figure}[]
	\centering
	\includegraphics[width=8cm]{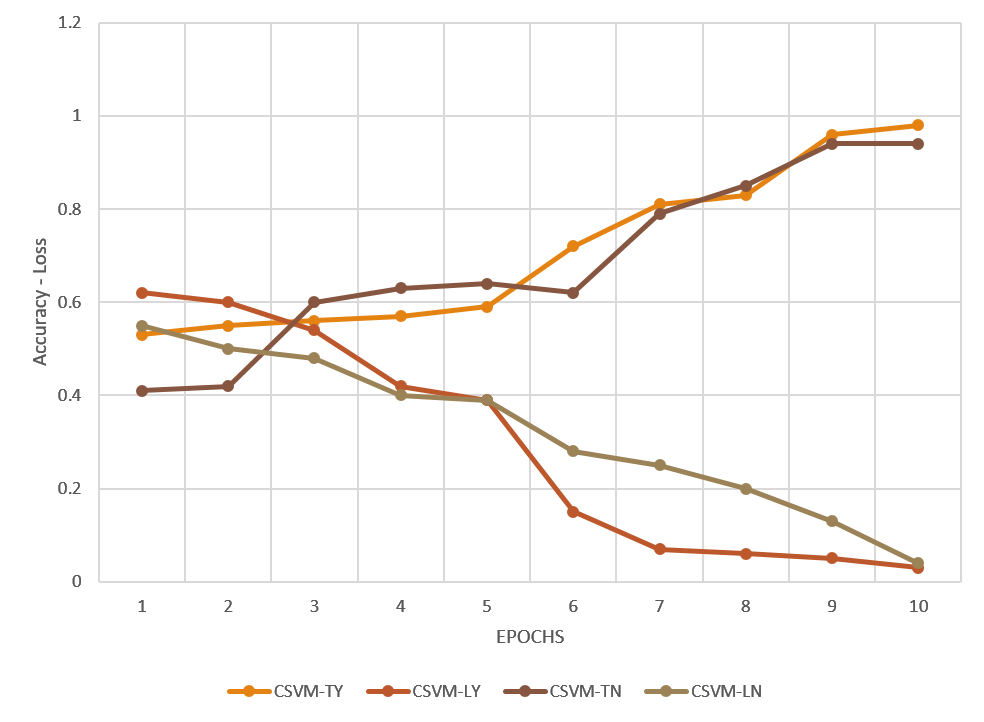}
	\caption{CSVM: Accuracy versus training loss with and without proposed preprocessing.}
	\label{fig:csvm_pp}
\end{figure}

\begin{figure}[]
  \centering
  \includegraphics[width=8cm]{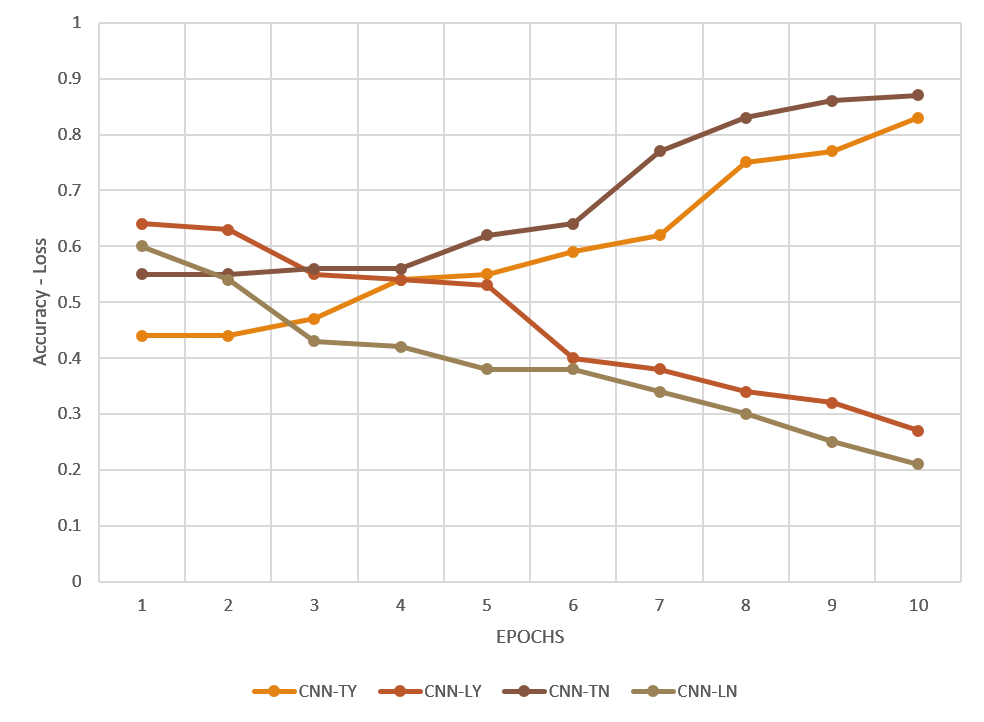}
  \caption{CNN: Accuracy versus training loss with and without proposed preprocessing.}
  \label{fig:cnn_pp}
\end{figure}

\textbf{\textit{Training evaluation:}} A loss function is used to optimize the machine learning algorithm by calculating the loss on the training MRI dataset images. Its interpretation is based on how well the model is performing for the dataset. It is the sum of errors made for each data points in training. This value implies how well or poor a model can perform after each iteration of optimization also known as epochs. On the other hand, the training accuracy metric is used to measure the algorithm’s performance based on the total number of correctly classified data instances in terms of a probability value during the training. For any model, our target is always to minimize the loss while still having good accuracy. To check the benefit of the proposed preprocessing technique over the models, four models are considered for the experiment. The four models compared in this regard are the proposed CSVM model, CNN, SVM, and Random Forest (RF). These models are chosen due to their popularity in this field of research and these models tend to provide decent results for image classification tasks. To test the effectiveness of the proposed preprocessing technique, the models are trained on the same dataset but one set with the proposed preprocessing applied and the other with the application of only basic preprocessing steps. Also, it is important to keep a note on the fact that the feature selection process in Section-\ref{S:3.3} only applies to CSVM and not other models. Other models use the general image features from the preprocessing step.

\begin{figure}[]
  \centering
  \includegraphics[width=8cm]{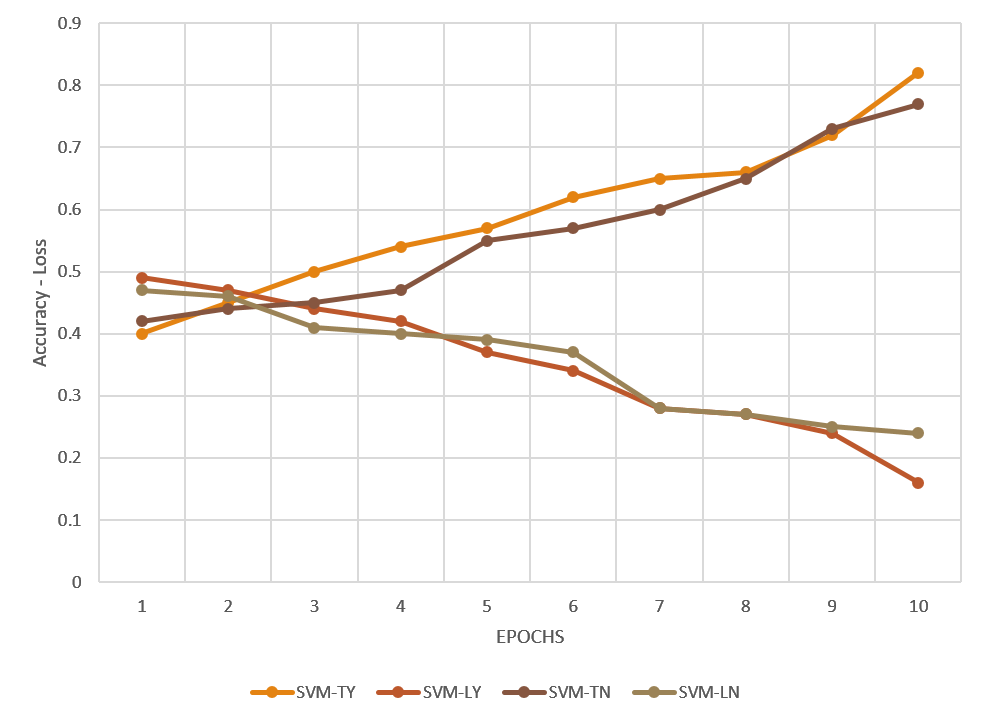}
  \caption{SVM: Accuracy versus training loss with and without proposed preprocessing.}
  \label{fig:svm_pp}
\end{figure}

\begin{figure}[H]
  \centering
  \includegraphics[width=8cm]{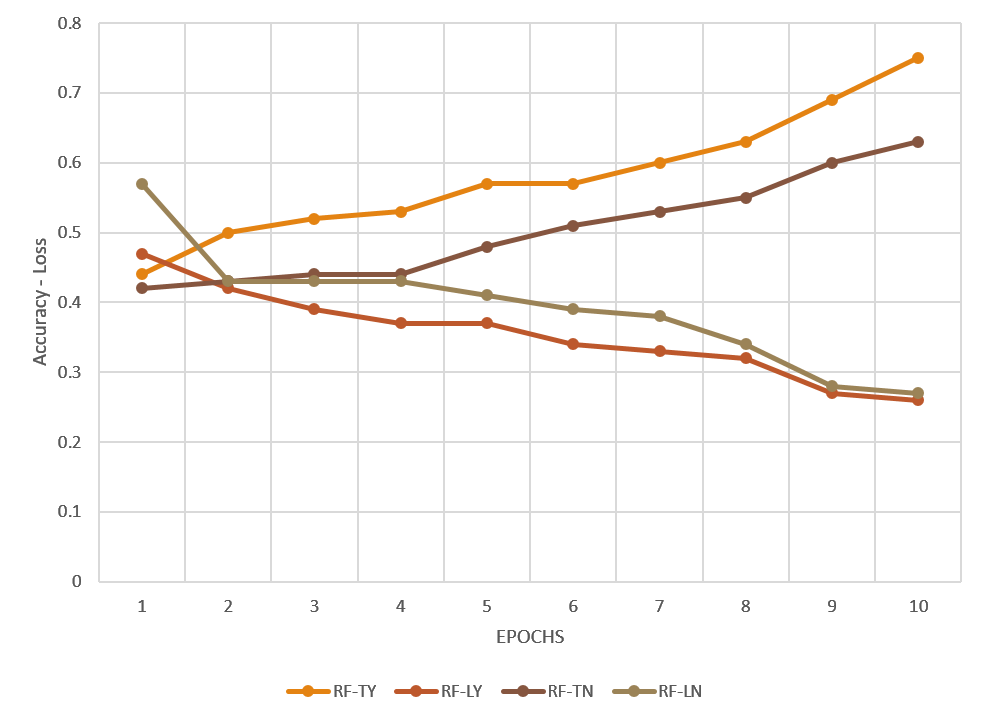}
  \caption{RF:Accuracy versus training loss with and without proposed preprocessing.}
  \label{fig:rf_pp}
\end{figure}

From the experimental results shown in Fig. \ref{fig:csvm_pp}, \ref{fig:cnn_pp}, \ref{fig:svm_pp}, \ref{fig:rf_pp}, it is observed that the accuracy for training when using proposed preprocessing is significantly better in most of the epochs compared to the same setting without the proposed preprocessing step. As per the training loss is concerned, the loss gets minimized much more with every epoch when the proposed preprocessing is applied. In some of the epochs, the accuracy or loss may be close but the final training accuracy with proposed preprocessing is higher and the loss is lower. The superiority of the hybrid CSVM model can also be seen in Fig. \ref{fig:csvm_pp}, as its training accuracy is consistently better than the other models. Further, the introduction of the advanced proposed preprocessing technique with all the models proves to be beneficial both in maximizing the accuracy and minimizing the loss. The cause of the poor performance of the CNN model is probably related to the very low number of items in the dataset. Neural Network models perform better for a large collection of data, in this case, prediction via hybrid CSVM proves the capability of SVM in dealing with low data count but still producing competitive results. For SVM without convolutional feature extraction, the training accuracy is decreased. Random Forest performed worst in this case with the lowest training accuracy. A more detailed comparison of the models is performed for the test data in the later section. For the rest of the experiment, the proposed preprocessing technique is considered regardless of the models used.

\subsubsection{Effect of the proposed cost optimization}
\label{S:4.4.2}

From the experimental model of CSVM, confusion matrices are generated for each case of classification, that is, for detecting the presence of neoplasm and for classifying the type of neoplasm. An example confusion matrix is shown in Figure-\ref{fig:cmatrix}. Confusion matrices by themselves do not provide much insight into how a model performs. But rater, confusion matrices can provide a fast at-a-glance insight about the classification. Further, lots of performance metrics, like accuracy, precision, recall, F-measure is derived from the confusion matrices. These metrics in turn provides a better understanding of the model. For performance evaluation, the metrics used are provided in Eqn. \ref{EquationA}, \ref{EquationP}, \ref{EquationR}, \ref{EquationF}.

\begin{equation}
    Accuracy(A) = \frac{TP+TN}{TP+TN+FP+FN}
    \label{EquationA}
\end{equation}

\begin{equation}
    Precision(P) = \frac{TP}{TP+FP}
    \label{EquationP}
\end{equation}

\begin{equation}
    Recall(R) = \frac{TP}{TP+FN}
    \label{EquationR}
\end{equation}

\begin{equation}
    F1-Score(F1) = 2\times \frac{P \times R}{P+R}
    \label{EquationF}
\end{equation}

\vspace{0.8cm}
where:\\
True Positives(TP): Correct classification of presence of neoplasm/malignant,\\
True Negatives(TN): Correct classification of absence of neoplasm/benign,\\
False Positives(FP): Incorrect classification of absence of neoplasm/benign,\\
True Negatives(TN): Incorrect classification of presence of neoplasm/malignant.\\\\

\begin{figure}[H]
	\centering
	\includegraphics[width=10cm, height=!]{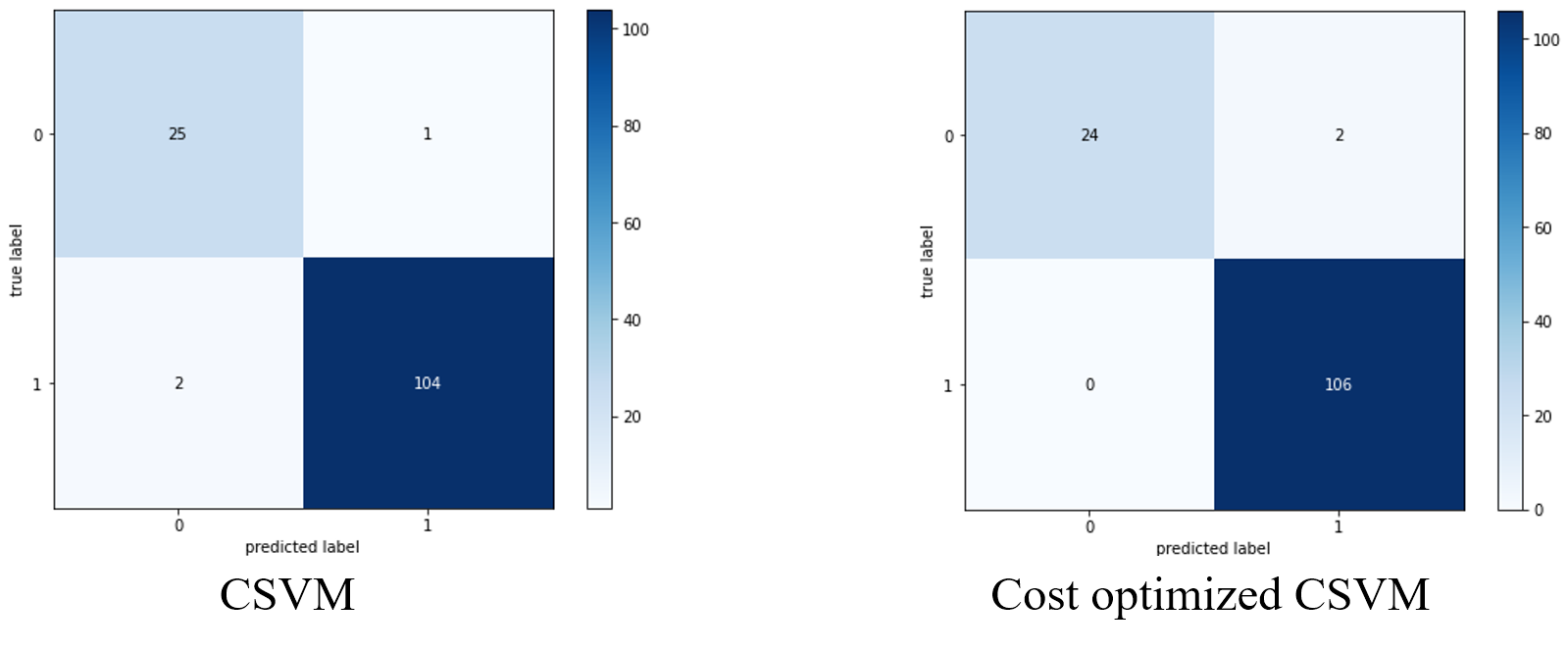}
	\caption{Confusion matrix for detecting presence of brain neoplasm.}
	\label{fig:cop}
\end{figure}

\vspace{-0.5cm}
For detecting the presence of neoplasm, a false negative detection would mean, some MRIs were detected as an absence of neoplasm when in fact there are neoplasms present. So, there shall be no diagnosis involved, and could be fatal. Whereas, false positive case there may be over-diagnosis but will not be fatal. So, we shall try to analyze the results based on these conditions. The confusion matrices for the detection of neoplasm in an MRI for CSVM approaches with and without cost optimization are represented in Fig. \ref{fig:cop}. From the confusion matrices, it is clear that the total false classifications of CSVM with and without cost optimization are 3 and 2 respectively. Further, as already discussed false positives are more tolerable than false negatives in this situation. There are 2 false negative classifications via the CSVM model without cost optimization and no false negative classifications in the case of CSVM with cost optimization. It shows the advantage of the CSVM model with cost optimization in this regard.

\vspace{-0.8cm}

\begin{figure}[ht]
	\centering
	\includegraphics[width=10cm, height=!]{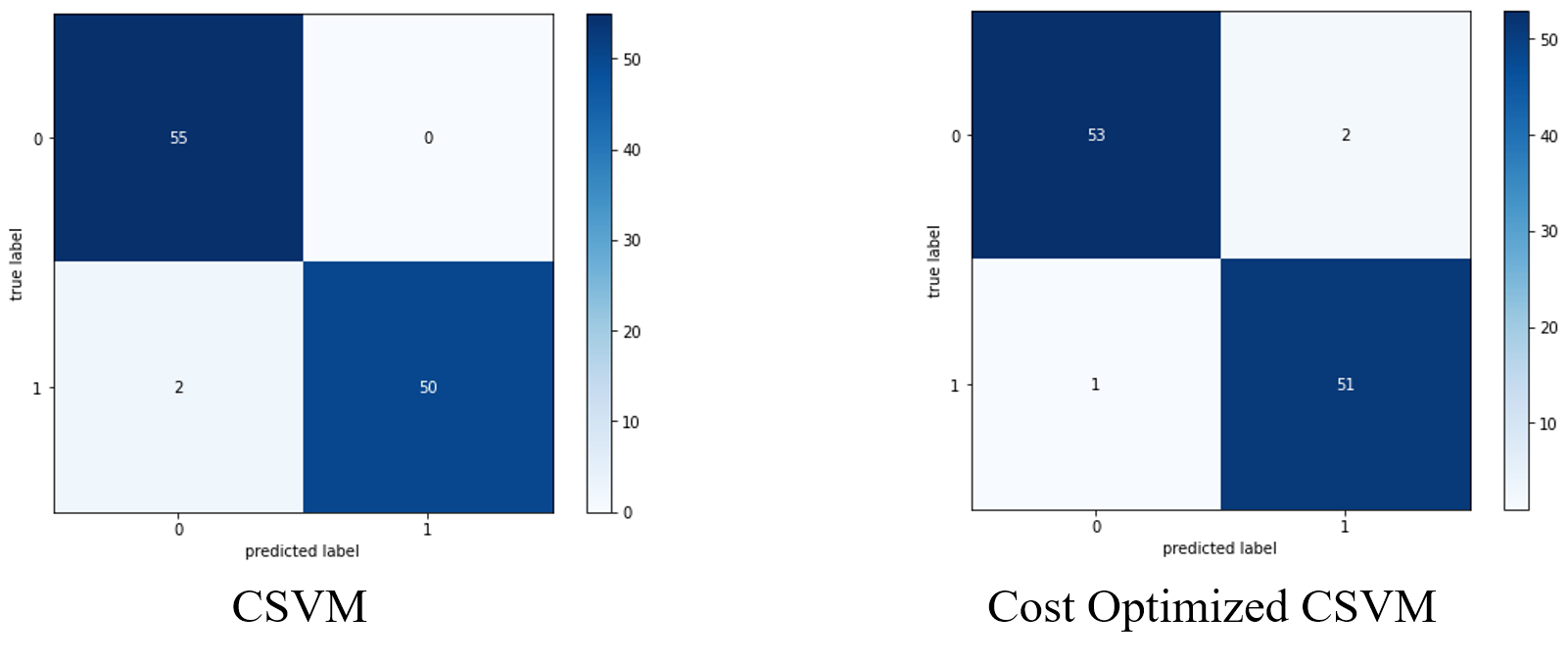}
	\caption{Confusion Matrix for Detecting Severity of Neoplasm}
	\label{fig:cos}
\end{figure}

For detecting the severity of neoplasm, a false negative detection would mean, some MRIs were detected as benign when in fact it is malignant. In this case, if the MRI is not re-evaluated quickly, there is a chance of fatality. False positives in this case may suffer from sock but are still more acceptable than the previous one. The confusion matrices for the detection of the severity of the neoplasm in an MRI using CSVM approaches are represented in Fig. \ref{fig:cos}. From the confusion matrices, it is clear that the total false classifications of CSVM with and without cost optimization are 2 and 3 respectively. There are 2 false negative classifications via the CSVM model without cost optimization and 1 false negative classification in the case of CSVM with cost optimization. It again shows the advantage of cost optimized CSVM model over the regular CSVM model for medical domain-specific data.

\subsubsection{Classification accuracy}
\label{S:4.4.3}

The MRI dataset of human brain images is split into 80:20 ratio for the training and testing of the models and then the experiments are performed. The experimental overview with the evaluation scores for detecting the presence of neoplasms using CSVM, CNN, SVM, and Random Forest are presented in Table \ref{tab:metrics_presence}. The training and testing columns have normal and neoplasm data points defined as N and B+M. As seen from the result, CSVM achieved maximum scores in all the aspects for detecting the presence of neoplasm.

\begin{table}[H]
\centering
\resizebox{\columnwidth}{!}{%
\begin{tabular}{c|c|c|c|c|c|cccc}
\cline{3-6}
\multicolumn{2}{l|}{} & \multicolumn{4}{c|}{{\color[HTML]{000000} \textbf{Number of MRI Images}}} & \multicolumn{4}{l}{{\color[HTML]{000000} \textbf{}}} \\ \cline{3-10} 
\multicolumn{2}{l|}{\multirow{-2}{*}{}} & \multicolumn{2}{c|}{\textit{Training}} & \multicolumn{2}{c|}{\textit{Testing}} & \multicolumn{4}{c|}{\textbf{Evaluation Metrics}} \\ \hline
\multicolumn{1}{|c|}{\textbf{Classifier}} & \textbf{Total Data} & \textit{N} & \textit{N+B} & \textit{N} & \textit{N+B} & \multicolumn{1}{c|}{\textit{Accuracy}} & \multicolumn{1}{c|}{\textit{Precision}} & \multicolumn{1}{c|}{\textit{Recall}} & \multicolumn{1}{c|}{\textit{F1-Score}} \\ \hline
\multicolumn{1}{|c|}{CSVM} & 658 & 104 & 422 & 26 & 106 & \multicolumn{1}{c|}{0.9848} & \multicolumn{1}{c|}{0.9815} & \multicolumn{1}{c|}{1} & \multicolumn{1}{c|}{0.9907} \\ \hline
\multicolumn{1}{|c|}{CNN} & 658 & 104 & 422 & 26 & 106 & \multicolumn{1}{c|}{0.9242} & \multicolumn{1}{c|}{0.9615} & \multicolumn{1}{c|}{0.9434} & \multicolumn{1}{c|}{0.9524} \\ \hline
\multicolumn{1}{|c|}{SVM} & 658 & 104 & 422 & 26 & 106 & \multicolumn{1}{c|}{0.8181} & \multicolumn{1}{c|}{0.9271} & \multicolumn{1}{c|}{0.8396} & \multicolumn{1}{c|}{0.8812} \\ \hline
\multicolumn{1}{|c|}{RF} & 658 & 104 & 422 & 26 & 106 & \multicolumn{1}{c|}{0.6894} & \multicolumn{1}{c|}{0.8652} & \multicolumn{1}{c|}{0.7264} & \multicolumn{1}{c|}{0.7898} \\ \hline
\end{tabular}%
}
\caption{\label{tab:metrics_presence}Evaluation scores for detecting presence of neoplasms.}
\end{table}

The evaluation scores for detecting the severity of neoplasms are presented in Table \ref{tab:metrics_severity}. The training and testing columns have benign and malignant data points defined as B and M in the table. As seen from the result, CSVM again achieved maximum scores in all the aspects for severity classification of brain neoplasm. Further, cost optimization in the CSVM model has shown significant improvement in recall score than other models.
\vspace{-0.4cm}
\begin{table}[H]
\centering
\resizebox{\columnwidth}{!}{%
\begin{tabular}{c|c|c|c|c|c|cccc}
\cline{3-6}
\multicolumn{2}{l|}{} & \multicolumn{4}{c|}{{\color[HTML]{000000} \textbf{Number of MRI Images}}} & \multicolumn{4}{l}{{\color[HTML]{000000} \textbf{}}} \\ \cline{3-10} 
\multicolumn{2}{l|}{\multirow{-2}{*}{}} & \multicolumn{2}{c|}{\textit{Training}} & \multicolumn{2}{c|}{\textit{Testing}} & \multicolumn{4}{c|}{\textbf{Evaluation Metrics}} \\ \hline
\multicolumn{1}{|c|}{\textbf{Classifier}} & \textbf{Total Data} & \textit{B} & \textit{M} & \textit{B} & \textit{M} & \multicolumn{1}{c|}{\textit{Accuracy}} & \multicolumn{1}{c|}{\textit{Precision}} & \multicolumn{1}{c|}{\textit{Recall}} & \multicolumn{1}{c|}{\textit{F1-Score}} \\ \hline
\multicolumn{1}{|c|}{CSVM} & 528 & 216 & 205 & 55 & 52 & \multicolumn{1}{c|}{0.972} & \multicolumn{1}{c|}{0.9623} & \multicolumn{1}{c|}{0.9808} & \multicolumn{1}{c|}{0.9714} \\ \hline
\multicolumn{1}{|c|}{CNN} & 528 & 216 & 205 & 55 & 52 & \multicolumn{1}{c|}{0.9346} & \multicolumn{1}{c|}{0.9412} & \multicolumn{1}{c|}{0.9231} & \multicolumn{1}{c|}{0.932} \\ \hline
\multicolumn{1}{|c|}{SVM} & 528 & 216 & 205 & 55 & 52 & \multicolumn{1}{c|}{0.8598} & \multicolumn{1}{c|}{0.8491} & \multicolumn{1}{c|}{0.8654} & \multicolumn{1}{c|}{0.8571} \\ \hline
\multicolumn{1}{|c|}{RF} & 528 & 216 & 205 & 55 & 52 & \multicolumn{1}{c|}{0.7477} & \multicolumn{1}{c|}{0.7451} & \multicolumn{1}{c|}{0.7308} & \multicolumn{1}{c|}{0.7379} \\ \hline
\end{tabular}%
}
\caption{\label{tab:metrics_severity}Evaluation scores for severity classification.}
\end{table}

\subsubsection{Comparison with the existing approaches}
\label{S:4.4.4}

To further evaluate the effectiveness of the CSVM approach, the final result is compared with other procedures using the same MRI dataset. The compared approaches omit the proposed preprocessing step and do not consider unequal classification costs associated with the MRI Brain neoplasm classification. The compared models use various techniques like Wavelet Transformation(WT), Spatial Gray Level Dependence Method (SGLDM), Genetic Algorithm (GA), and Principle Component Analysis(PCA). The result is shown in Table \ref{tab:compare_sota}. From the comparison, it is seen that CSVM performed best for both the classification task i.e. neoplasm presence detection and severity classification from the compared models. Detection of the presence of neoplasm is having a 15.26\% improvement over the compared technique. For severity detection, the CSVM model is 1\% and 3\% more accurate than the compared procedures. As already discussed, maximization of recall is necessary for this domain, we can see a difference of 3.5\% and 6.2\% respectively from the compared procedures with CSVM.

\begin{table}[H]
\centering
\resizebox{\columnwidth}{!}{%
\begin{tabular}{c|c|c|c|c|c|}
\cline{3-6}
\multicolumn{2}{l|}{} & \multicolumn{4}{c|}{{\color[HTML]{000000} Evaluation Metrics}} \\ \hline
\multicolumn{1}{|c|}{Classifier} & Classification Task & Accuracy & Precision & Recall & F1-Score \\ \hline
\multicolumn{1}{|c|}{CSVM} & \textit{Detect Neoplasm Presence} & 0.9848 & 0.9815 & 1 & 0.9907 \\ \hline
\multicolumn{1}{|c|}{CSVM} & \textit{Detect   Neoplasm Severity} & 0.9720 & 0.9623 & 0.9808 & 0.9714 \\ \hline
\multicolumn{1}{|c|}{\begin{tabular}[c]{@{}c@{}}WT+SGLDM\\    +GA+SVM\end{tabular}} & \textit{Detect   Neoplasm Severity} & 0.9629 & N/A & 0.9460 & N/A \\ \hline
\multicolumn{1}{|c|}{\begin{tabular}[c]{@{}c@{}}SGLDM+GA   \\ +SVM\end{tabular}} & \textit{Detect   Neoplasm Severity} & 0.9444 & N/A & 0.9187 & N/A \\ \hline
\multicolumn{1}{|c|}{\begin{tabular}[c]{@{}c@{}}PCA+GA\\  +SVM\end{tabular}} & \textit{Detect Neoplasm Presence} & 0.8322 & N/A & N/A & N/A \\ \hline
\end{tabular}%
}
\caption{\label{tab:compare_sota}Comparison of proposed model with other procedures.}
\end{table}

\section*{Conclusions}
A hybrid preprocessing using contour plot and Sobel edge is proposed and results are observed by including and excluding this preprocessing step. The results signify the benefit of using the proposed preprocessing technique, as the model performance is increased in each case. Further, the proposed CSVM performed better than a regular CNN, SVM, or Random Forest model every time in classifying between normal and abnormal brain MRI and also for benign and malignant. Cost optimization also helped in improving model performance and reduced false negative predictions. Comparison with other procedures using the same dataset also proves the efficacy of the overall proposed techniques.

\bibliographystyle{spmpsci}
\bibliography{ref}

\vskip5pt

\section*{Authors' Biographies}

{
\begin{wrapfigure}{l}{25mm} 
    \includegraphics[width=1in,height=1.25in,clip,keepaspectratio]{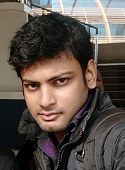}
  \end{wrapfigure}\par \textbf{Nilanjan Sinhababu}  received his B. Tech. degree in Computer Science \& Engineering from West Bengal Technical University, India. He is currently pursuing his Master of Science (by research) degree Subir Chwdhury School of Quality and Reliability, Indian Institute of Technology Kharagpur. His research interests include recommendation system, data analytics, Machine learning, deep learning, artificial intelligence, etc.  

}
\vspace{1cm}
{
\begin{wrapfigure}{l}{25mm} 
    \includegraphics[width=1in,height=1.25in,clip,keepaspectratio]{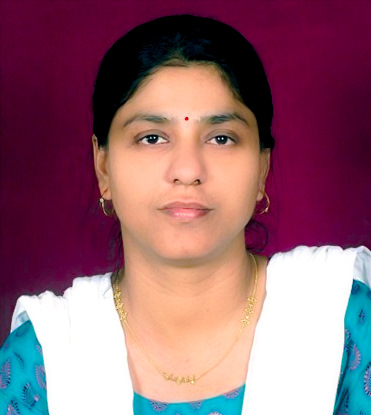}
  \end{wrapfigure} \textbf{Monalisa Sarma} received her Ph.D. degree in Computer Science \& Engineering from Indian Institute of Technology Kharagpur, India. She holds M.S. (by research) and B. Tech. degrees both in Computer Science \& Engineering from Indian Institute of Technology Kharagpur, India, and North Hill University, India, respectively. Presently, she is an assistant professor, Reliability Engineering Centre, India Institute of Technology Kharagpur. Prior to joining Indian Institute of Technology Kharagpur, she was working in the Department of Computer Science \& Engineering, Indian Institute of Technology Indore and Siemens Research and Devolvement, Bangalore, India. Her current research includes human reliability, big data security, biometric-based cryptography, etc. \par

}
\vspace{1cm}
{
\begin{wrapfigure}{l}{25mm} 
    \includegraphics[width=1in,height=1.25in,clip,keepaspectratio]{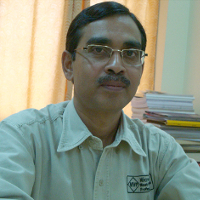}
  \end{wrapfigure}\par\textbf{Debasis Samanta} received his Ph.D. degree in Computer Science \& Engineering from Indian Institute of Technology Kharagpur, India. He holds M. Tech. and B. Tech. degrees both in Computer Science \& Engineering from Jadavpur University, Kolkata, India and Calcutta University, India, respectively. Presently, he is an Associate Professor, Department of Computer Science \& Engineering, Indian Institute of Technology Kharagpur. His current research includes Human Computer Interaction, Brain Computing Interaction, Biometric-based System Security, and Data Analytics.\par
}

\end{document}